\documentclass[preprint,12pt]{elsarticle}

\usepackage{amssymb}
\usepackage{graphicx}
\usepackage{epsfig}
\usepackage{graphicx}
\usepackage{amssymb}
\usepackage{longtable}

\usepackage{mathrsfs}
\usepackage{subcaption}
\usepackage{graphicx}
\usepackage{mathtools}
\usepackage{amssymb}
\usepackage{color}
\usepackage{lineno}
\usepackage{graphicx}
\usepackage{amssymb}
\usepackage{amsmath}
\usepackage{makecell}
\usepackage{algorithm} 
\usepackage{algpseudocode} 
\usepackage{pdflscape}
\usepackage{adjustbox}
\usepackage[utf8]{inputenc}
\usepackage{adjustbox}
\usepackage{tabularx}
\usepackage{blindtext}
\usepackage{setspace}
\usepackage{graphicx}
\usepackage{notoccite} 
\usepackage{lscape} 
\usepackage{caption} 
\usepackage{longtable}
\usepackage{pdflscape}
\usepackage{mwe}
\usepackage{pifont}
\usepackage{caption}
\usepackage{subcaption}
\usepackage{adjustbox}
\usepackage{tabularx}
\usepackage{booktabs}
\usepackage{makecell}
\usepackage{graphicx}
\usepackage{longtable}
\restylefloat{table}
\usepackage{xcolor, soul}
\usepackage{graphicx}
\usepackage{epstopdf}

\journal{Elsevier}

\begin{document}
\begin{frontmatter}



\title{Projection based fuzzy least squares twin support vector machine for class imbalance problems}


\author[inst1]{M. Tanveer\corref{mycorrespondingauthor}}
\author[inst1]{Ritik Mishra}
\affiliation[inst1]{organization={Department of Mathematics, Indian Institute of Technology Indore, Simrol, Indore 452552},
            country={India}}
            \cortext[mycorrespondingauthor]{Corresponding author}

\author[inst2]{Bharat Richhariya}

\affiliation[inst2]{organization={Department of Computer Science $\&$ Information Systems, BITS Pilani (Pilani Campus)},
            country={India}}

\begin{abstract}
Class imbalance is a major problem in many real world classification tasks. Due to the imbalance in the number of samples, the support vector machine (SVM) classifier gets biased toward the majority class. Furthermore, these samples are often observed with a certain degree of noise. Therefore, to remove these problems we propose a novel fuzzy based approach to deal with class imbalanced as well noisy datasets. We propose two approaches to address these problems. The first approach is based on the intuitionistic fuzzy membership, termed as robust energy-based intuitionistic fuzzy least squares twin support vector machine (IF-RELSTSVM). Furthermore, we introduce the concept of hyperplane-based fuzzy membership in our second approach, where the final classifier is termed as robust energy-based fuzzy least square twin support vector machine (F-RELSTSVM). By using this technique, the membership values are based on a projection based approach, where the data points are projected on the hyperplanes. The performance of the proposed algorithms is evaluated on several benchmark and synthetic datasets. The experimental results show that the proposed IF-RELSTSVM and F-RELSTSVM models outperform the baseline algorithms. Statistical tests are performed to check the significance of the proposed algorithms. The results show the applicability of the proposed algorithms on noisy as well as imbalanced datasets.
\end{abstract}



\end{frontmatter}


\section{Introduction}
Support vector machines (SVMs) \cite{cortesc1995support} are computationally powerful tools for supervised learning, and therefore frequently utilized in classification \cite{tanveer2020machine} and regression problems \cite{brereton2010support}. SVMs have been effectively used to solve a range of real world problems, including intrusion detection system \cite{khan2007new}, facial expression recognition \cite{richhariya2019facial}, speaker identification \cite{schmidt1996speaker}, text classification \cite{zhang2006class}, and seizure detection \cite{richhariya2018eeg}. The methodical approach of SVMs is driven by the statistical learning theory. Finding the best-separating hyperplane between the positive and negative instances is the main goal of SVM.
However, despite their effectiveness, SVMs have some limitations, such as higher time complexity, sensitivity to noise and outliers, and inability to handle class imbalance problems \cite{batuwita2010fsvm}.
\par To address the limitations mentioned above, several variants of SVM have been proposed, such as generalized eigenvalue problem-based SVM (GEPSVM) \cite{sun2018multiview} and fuzzy support vector machine (FSVM) \cite{khemchandani2007twin}. GEPSVM solves a generalized eigenvalue problem which is computationally efficient and provides a better solution than SVM. Jayadeva et al. \cite{khemchandani2007twin} proposed a twin support vector machine (TWSVM) that constructs two non-parallel hyperplanes like GEPSVM. However, unlike GEPSVM, TWSVM solves a pair of smaller-sized quadratic programming problems (QPPs) instead of a generalized eigenvalue problem. TWSVM is efficient with respect to time complexity in comparison to SVM. To reduce the computational complexity of TWSVM, the least squares twin support vector machine (LSTSVM) \cite{ARUNKUMAR20097535} was proposed as an alternative to the convex QPPs in TWSVM. LSTSVM utilizes the squared loss function instead of the hinge loss function, resulting in lower training time \cite{wang2020comprehensive}. However, the LSTSVM model needs the hyperplane at exactly one distance from the other class, making it sensitive to noise and outliers.
\par To address the sensitivity to noise near the hyperplanes, Nasiri et al. \cite{nasiri2014energy} developed the energy-based LSTSVM (ELS-TSVM) model, which introduces energy parameters to relax the constraints for the hyperplanes. Furthermore, Tanveer et al. \cite{tanveer2016robust} developed a robust energy-based least square twin support vector machine (RELS-TSVM) by incorporating regularization terms in ELS-TSVM, making it more robust to noise.
Recent studies have shown that the RELS-TSVM model and its variants outperform other TWSVM-based models in binary class problems because they can handle noise and outliers effectively \cite{TANVEER2019105617}. Recently, Laxmi et al. \cite{laxmi2022intuitionistic} suggested another TWSVM variant, intuitionistic fuzzy least square twin support vector machine (IFLSTSVM). IFLSTSVM incorporates the concept of the intuitionistic fuzzy set \cite{atanassov1999intuitionistic}, which handles the uncertainties in the data more precisely.
\par One of the frequently encountered problems in real world classification tasks is class imbalance \cite{kaur2019systematic,richhariya2018robust,richhariya2020reduced}. In case of RELS-TSVM, the classifier gives equal weightage to each sample, causing the learned decision surface to get biased toward the majority class. To address this issue, we propose two improved variants of RELS-TSVM algorithm, known as robust energy-based intuitionistic fuzzy least squares twin support vector machine (IF-RELSTSVM) and robust energy-based fuzzy least squares twin support vector machine (F-RELSTSVM) for class imbalance learning. Unlike TWSVM, LSTSVM, ELS-TSVM, and RELS-TSVM models, the proposed IF-RELSTSVM introduces intuitionistic fuzzy scores to both classes to reduce the negative effect of noise and outliers. This intuitionistic fuzzy score is achieved by introducing a pair of membership and non-membership degrees for every data point. The proposed IF-RELSTSVM model also incorporates a regularization term to reduce over-fitting. This extra component encapsulates the marrow of statistical learning theory and implements the structural risk minimization (SRM) principle in the proposed formulation.

Furthermore, in the proposed F-RELSTSVM model, we present another approach where the projection on proximal hyperplanes is utilized for the fuzzy memberships.
Both the proposed IF-RELSTSVM and F-RELSTSVM models solve systems of linear equations instead of QPPs as in TWSVM. Therefore, the proposed models are robust compared to the TWSVM, LSTSVM, and ELS-TSVM. Moreover, the proposed models can handle the class imbalance problem by introducing a weight parameter to the objective function, which balances the influence of the minority and majority classes in the learning process. To evaluate the performance of the proposed algorithms, we conducted experiments on several benchmark datasets. Experimental results show that the performance of the proposed F-RELSTSVM model is better than other state-of-the-art algorithms, including LSTSVM, ELS-TSVM, RELS-TSVM, and IFLSTSVM in terms of AUC \cite{huang2005using}.
\par The proposed IF-RELSTSVM and F-RELSTSVM models have the following appealing features:
\begin{enumerate}
\item Unlike TWSVM, LSTSVM, ELS-TSVM, and RELS-TSVM, the proposed IF-RELSTSVM model introduces intuitionistic fuzzy score to both classes that reduce the effect of noise and outliers in samples. IF-RELSTSVM introduces a pair of membership and non-membership degrees for every data point. Moreover, the membership degree uses the imbalance ratio (IR) as a multiplier to deal with class imbalance issues in the datasets. So, the problems of class imbalance and noisy data are handled in this approach.

\item The proposed IF-RELSTSVM algorithm provides solution to two types of noise:
\begin{enumerate}
\item  Noise near the hyperplane which is dealt with energy-based approach. 
\item  Noise away from the hyperplane which is dealt with the intuitionistic fuzzy-based approach in this work.
\end{enumerate}

\item The proposed F-RELSTSVM model utilizes the projection on proximal hyperplanes to measure the membership leading to proper fuzzy memberships to the data points before formulating the final classifier.

\item IF-RELSTSVM and F-RELSTSVM involve the regularisation term to each objective function to maximize the margin. This involves the structural risk minimization principle in the proposed formulations.

\item The proposed formulations involve the solution of system of linear equations instead of QPPs, leading to lesser computation time.

\end{enumerate}

\section{Related work}
This section discusses the intuitionistic fuzzy set and formulations of LSTSVM, ELS-TSVM, and RELS-TSVM algorithms.

Suppose $X = \lbrace(x_1,y_1),(x_2,y_2),...,(x_m,y_m) \rbrace$ is a set of training samples, where $x_l \in \mathbb{R}^n$ being the $l^{th}$ training sample, $l \in \{1,2,...,m\}$ and $y_i \in \{-1,+1\}$ being the associated target class.
\subsection{Intuitionistic fuzzy set}
Let $\mathrm{A}$ be a nonempty set. A fuzzy set \cite{ZADEH1965338} $F$ in the universe $\mathrm{A}$ is defined by $$F=\left\{\left(y, \mu_F(y)\right): y \in A\right\},$$ where, $\mu_F: X \rightarrow[0,1]$ is the membership (degree of belongingness) of $y$ to $A$.\\
Intuitionistic fuzzy ($IF$) set  \cite{atanassov1999intuitionistic} can be defined as $$I F=\left\{\left(y, \mu_{I F}(y), \eta_{I F}(y)\right): y \in A\right\},$$
where $\mu_{I F}: X \rightarrow[0,1], \eta_{I F}: X \rightarrow[0,1], 0 \leq \mu_{I F}(y)+\eta_{I F}(y) \leq 1, \mu_{I F}$ $ \text{and} $ $ \eta_{I F}$ represent the membership and non-membership degrees of belongingness of $y$ in $A$, respectively.\\
For a data point $y$, the degree for not belonging to $A$ is defined by $$h_{I F}(y)=1-\mu_{I F}(y)-\eta_{I F}(y).$$
An intuitionistic fuzzy number ($IFN$) is defined by $$IFN=\left(\mu_{I F}, \eta_{I F}\right).$$
Hence, $(1,0)$ represents the highest membership value and least non-membership value, while $(0,1)$ represents the least membership value and the highest non-membership value.
Therefore, the largest intuitionistic fuzzy number and smallest intuitionistic fuzzy numbers are $(1,0)$ and $(0,1)$, respectively. 
For a given $IFN$ the score value is given as $$\mathrm{s}({IFN})=\mu_{I F N}-\eta_{I F N}.$$
Similarly, another precise score value \cite{ha2013support} can be defined as 
$$
\begin{aligned}
& p(IFN)=\mu_{I F N}+\eta_{I F N}, \\
& \therefore p(I F N)+h_{I F}(I F N)=1 .
\end{aligned}
$$
For given $I F N_1$ and $I F N_2$, we can say that $I F N_1<I F N_2$ if
$$
s\left(I F N_1\right)=s\left(I F N_2\right) \\ \text { and } p\left(I F N_1\right)<p\left(I F N_2\right).
$$
Here, we define a derived score using the precise score values as
$$
K(I F N)=\frac{1-\eta_{I F N}}{2-\mu_{I F N}-\eta_{I F N}}.
$$
Thus, there is conjunction between membership and non-membership values as follows:
\begin{enumerate}
    \item $s\left(I F N_1\right)<s\left(I F N_2\right) \Rightarrow K\left(I F N_1\right)<K\left(I F N_2\right)$,
    \item $s\left(I F N_1\right)=s\left(I F N_2\right),$ $p\left(I F N_1\right)<p\left(I F N_2\right) \Rightarrow K\left(I F N_1\right)<K\left(I F N_2\right)$.
\end{enumerate}
\subsection{Least squares twin support vector machine (LSTSVM)}
To minimize the computational complexity of TWSVM, Kumar and Gopal \cite{ARUNKUMAR20097535}  introduced a least squares twin support vector machine (LSTSVM). LSTSVM involves solving a pair of linear equations to obtain two hyperplanes. The formulation of LSTSVM is as follows:

For the binary classification problem, we consider two matrices $A$ and $B$, consisting of data points with class labels '+1' and '-1' respectively. The dimensions of matrices $A$ and $B$ are $p \times n$ and $q \times n$ respectively, where $n$ is the number of features. Then, the optimization problem of linear LSTSVM is formulated as follows:
\begin{align} \label{LS1}
&\underset{w_{1},b_{1},\xi _{2}}{\min }\,\frac{1}{2}\parallel Aw_{1}+e_{1}b_{1}\parallel ^{2}+C_{1}\xi_{2}^{T}\xi _{2} \nonumber \\ &\text{subject to}\ -(Bw_{1}+e_{2}b_{1})+\xi _{2} = e_{2},\, \end{align}
and 
\begin{align} \label{LS2}
&\underset{w_{2},b_{2},\xi _{1}}{\min }\,\frac{1}{2}\parallel Bw_{2}+e_{2}b_{2}\parallel ^{2}+C_{2}\xi_{1}^{T}\xi _{1} \nonumber \\ &\text{subject to}\ (Aw_{2}+e_{1}b_{2})+\xi _{1} = e_{1}.\  \end{align}


After putting the equality constraints in the objective function (\ref{LS1}), we get
\begin{align}\label{LS3}
\underset{w_{1},b_{1}}{\min }\,\frac{1}{2}\left\|A w_1+e_1 b_1\right\|^2+\frac{C_1}{2}\left\|B w_1+e_1 b_1+e_2\right\|^2.
\end{align}
Setting the gradient of (\ref{LS3}) with respect to $w_1$, $b_1$ and then equating it to zero, we get:
\begin{align}\label{LS4}
& A^T\left(A w_1+e_1 b_1\right)+C_1 B^T\left(B w_1+e_2 b_1+e_2\right)=0, \\ \label{LS5}
& e_1^T\left(A w_1+e_1 b_1\right)+C_1 e_2^T\left(B w_1+e_2 b_1+e_2\right)=0 .
\end{align}
Defining $G=\begin{bmatrix}A & e_1\end{bmatrix}$ and $H=\begin{bmatrix}B & e_2\end{bmatrix}$,
the solution becomes:
\begin{align} 
\label{LS7}
   \begin{bmatrix}
w_1 \\
b_1
\end{bmatrix}=-\left(H^TH+\frac{1}{C_1} G^T G\right)^{-1} H^T e_2. 
\end{align}
Similarly, the solution of QPP (\ref{LS2}) can be shown as,
\begin{align}\label{LS8}
\begin{bmatrix}
w_2 \\
b_2
\end{bmatrix}=\left(G^TG+\frac{1}{C_2} H^T H\right)^{-1} G^T e_1 . 
\end{align}
To decrease the computational time required for inverting matrices, the Sherman-Morrison-Woodbury (SMW) formula  \cite{golub1996matrix} is employed to solve equation (\ref{LS7}) and (\ref{LS8}) by finding the inverses of smaller dimension.
A new data point $x \in  \mathbb{R}^n$ is assigned
to a class using the following decision function: 
\begin{align}
   \text{class label}(x)=\arg\min_{i\in 1,2}\frac{\mid w_{i}^{T}x+b^{}_{i} \mid }{\parallel w^{ }_{i} \parallel }. 
\end{align} 
where $\mid.\mid$ represents the perpendicular distance between a data point and hyperplane.
\subsection{Energy-based least squares twin support vector machine (ELS-TSVM)}
To reduce the sensitivity of LSTSVM algorithm to noise and outliers, Nasiri et al. \cite{nasiri2014energy} proposed an energy-based least square twin support vector machine that incorporates an energy parameter for each hyperplane to limit the effect of noise and outliers. The optimization problem of ELS-TSVM is defined as:       
\begin{align} \label{EL1}
&\underset{w_{1},b_{1},\xi _{2}}{\min }\,\frac{1}{2}\parallel Aw_{1}+e_{1}b_{1}\parallel ^{2}+C_{1}\xi_{2}^{T}\xi _{2} \nonumber \\ &\text{subject to}\ -(Bw_{1}+e_{2}b_{1})+\xi _{2} = E_{2},\, \end{align}
and 
\begin{align} \label{EL2}
&\underset{w_{2},b_{2},\xi _{1}}{\min }\,\frac{1}{2}\parallel Bw_{2}+e_{2}b_{2}\parallel ^{2}+C_{2}\xi_{1}^{T}\xi _{1} \nonumber \\ &\text{subject to}\ (Aw_{2}+e_{1}b_{2})+\xi _{1} = E_{1},\  \end{align}
where $C_1$, $C_2$ are positive integers and $E_1$, $E_2$ are energy parameters.

On substituting the equality constraints into the objective function of QPP (\ref{EL1}), we get:
\begin{align}\label{EL3}
 L_1=\frac{1}{2}\left\|A w_1+e_1 b_1\right\|^2+\frac{C_1}{2}\left\|B w_1+e_2 b_1+E_1\right\|^2 .   
\end{align}
Setting the gradient of (\ref{EL3}) with respect to $w_1$ and $b_1$ and equating to zero, the solution of QPP (\ref{EL1}) is obtained as follows:
\begin{align}\label{EL4}
  \begin{bmatrix}
w_1 \\
b_1
\end{bmatrix}=-\left(C_1 H^T H+G^T G\right)^{-1} C_1 H^T E_1  ,
\end{align}
where $G=\begin{bmatrix}A & e_1\end{bmatrix}$ and $H=\begin{bmatrix}B & e_2\end{bmatrix}$.
Similarly, we get the solution of QPP (\ref{EL2}) as follows:
\begin{align}\label{EL5}
 \begin{bmatrix}
w_2 \\
b_2
\end{bmatrix}=\left(C_2 G^T G+H^T H\right)^{-1} C_2 G^T E_2 \text {. }   
\end{align}
A new sample $x$ is assigned depending on the following decision function:
\begin{align} \label{DF} \text{class label}(x)=\begin{cases}+1, &\frac{\mid w_{1}^{T}x+b_{1}\mid }{\mid w_{2}^{T}x+b_{2} \mid } \leq 1,\\ -1,&\frac{\mid w_{1}^{T}x+b_{1}\mid }{\mid w_{2}^{T}x+b_{2} \mid } > 1, \end{cases}  \end{align}

\subsubsection{ELS-TSVM vs LSTSVM algorithm}
In this section, we compare ELS-TSVM with the LSTSVM algorithm.
\begin{enumerate}
    \item The LSTSVM algorithm constraints demand the hyperplane to be exactly one distance from the points of the opposite class, making LSTSVM sensitive to outliers. For each hyperplane, ELS-TSVM includes an energy parameter that limits the impact of noise and outliers, and various energy values are chosen depending on past information or the grid search technique.
    \item The ELS-TSVM decision function is attained by using primal problems. However, the decision function in LSTSVM is computed using the perpendicular distance.
    \item The ELS-TSVM and LSTSVM algorithms are least squares variants of TWSVM that solve linear systems instead of solving the convex QPPs. As a result, ELS-TSVM and LSTSVM algorithms are much faster than TWSVM and have better generalization performance.
\end{enumerate}
\subsection{Robust energy-based least squares twin support vector machine (RELS-TSVM) }
By incorporating the regularization terms in the optimization of the ELS-TSVM algorithm, the linear case of RELS-TSVM \cite{tanveer2016robust} is formulated by the following pair of QPPs:
\begin{align} \label{REL1}&\underset{w_{1},b_{1},\xi _{2}}{\min }\,\frac{1}{2}\parallel Aw_{1}+e_{1}b_{1}\parallel ^{2}+C_{1}\xi_{2}^{T}\xi _{2} +\frac{1}{2}C_3(\parallel w_{1}\parallel ^{2}+ b_{1} ^{2})\nonumber \\ &\text{subject to}\ -(Bw_{1}+e_{2}b_{1})+\xi _{2} = E_{2},\, \end{align}
and 
\begin{align}\label{REL2} &\underset{w_{2},b_{2},\xi _{1}}{\min }\,\frac{1}{2}\parallel Bw_{2}+e_{2}b_{2}\parallel ^{2}+C_{2}\xi_{1}^{T}\xi _{1} +\frac{1}{2}C_4(\parallel w_{2}\parallel ^{2}+ b_{2}  
 ^{2})\nonumber \\ &\text{subject to}\ (Aw_{2}+e_{1}b_{2})+\xi _{1} = E_{1}. \end{align}
Substituting equality constraints into the objective function, we get the solution for QPP (\ref{REL1}).\\
Let 
$P=\begin{bmatrix}
   A & e_1 
\end{bmatrix}$ and  $Q=\begin{bmatrix}B & e_2\end{bmatrix}$. Then the solution of QPP (\ref{REL1}) is given as:  
\begin{align} \label{REL3} \left[\begin{array}{ll}w_1 \\ b_1\end{array}\right]=-\left(P^{T}P+C_{1}Q^{T}Q+C_{3}I\right)^{-1}C_{3}Q^{T}E_2 . \end{align}
In a similar way, the solution for QPP (\ref{REL2}) is given by 
\begin{align} \label{REL4}\left[\begin{array}{ll}w_2 \\ b_2\end{array}\right]=\left(Q^{T}Q+C_{2}P^{T}P+C_{4}I\right)^{-1}C_{3}P^{T} E_1 . \end{align}
To reduce the computational time required for inverting matrices, the SMW formula  \cite{golub1996matrix} is employed to solve equations (\ref{REL3}) and (\ref{REL4}). A new sample $x$ is assigned to a class using equation (\ref{DF}).
\subsection{Intuitionistic fuzzy membership assignment (IFMA)}
In intuitionistic fuzzy membership assignment \cite{ha2013support}, we employ a score value based on a pair of membership degrees and non-membership degrees to reduce the effect of noise and outliers. In IFMA, the membership degree is calculated in the feature space.

\subsubsection{Membership degree} 
\begin{figure}
\centering
  \includegraphics[width=3.5in,height=2.5in]{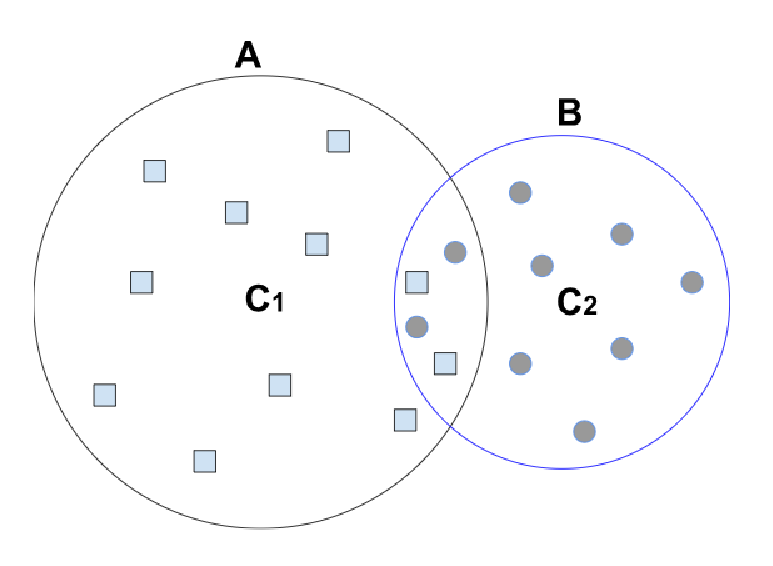}
\caption{Similar degree of membership for the training samples.}
\label{fig:IF}
\end{figure}

To determine the membership degree of the training data points,
we first project the training points
in the higher dimensional space. Then, we find the center of both classes and calculate
the distance for each training point from its class center to determine the membership
values. Given a training data point $x_i$ one can assign the membership value as:
\begin{align} \label{IF1}\mu (x_{i})=\begin{cases}1-\frac{\Vert \phi (x_{i})-C^{+}\Vert }{r^{+}+\delta }, & y_{i}=+1,\\1-\frac{\Vert \phi (x_{i})-C^{-}\Vert }{r^{-}+\delta }, & y_{i}=-1. \end{cases} 
\end{align}
In equation (\ref{IF1}), $C^{+} (C^{-})$ and $r^{+} (r^{-})$ are the center and radius of the class of the positive (negative) class $\delta >0$ is a small and adjustable parameter, , and
$\mid \mid.\mid \mid$ denotes the Frobenius norm.
\begin{align} \label{IF2} D(\phi (x_{i}),\phi (x_{j}))=\Vert \phi (x_{i})-\phi (x_{j})\Vert ,\end{align}
where $\phi$ maps the input data point into the higher dimensional space.
One can calculate the center of class as
\begin{align} \label{IF3} C^{\pm }=\frac{1}{l_{\pm }}\sum _{y_{i}=\pm 1}\phi (x_{i}) ,\end{align}
where $l_{\pm }$ is the cardinality of samples of positive (negative) class.\\
Also, we can calculate the radius of a class as
\begin{align} \label{IF44} r^{\pm }=\underset{y_{i}=\pm 1}\max \Vert \phi (x_{i})-C^{\pm }\Vert . \end{align}
It is visible in Fig. (\ref{fig:IF}) that the data points lying in the intersection region of the two classes have the same membership values with respect to both classes.
\subsubsection{Non-membership degree}
The degree of non-membership for a data point is given by the ratio of the number of heterogeneous points and all points in its neighborhood. The non-membership function is written as:
\begin{align} \label{IF4} \nu (x_{i})=(1-\mu (x_{i}))\rho (x_{i}) ,\end{align}
where $\rho (x_{i})$ can be calculated as
\begin{align} \label{IF5} \rho (x_{i})=\frac{|\lbrace x_{j}|\Vert \phi (x_{i})-\phi (x_{j})\Vert \leq \gamma,\,y_{j}\ne y_{i}\rbrace |}{|\lbrace x_{j}|\Vert \phi (x_{i})-\phi (x_{j})\Vert \leq \gamma \rbrace |}.\end{align}
In equation (\ref{IF5}), $\gamma > 0$ is an adjustable parameter and $|A|$ represents the cardinality of $A$.\\
The training points are transformed into intuitionistic fuzzy numbers,
\begin{eqnarray*} T= \Bigl\{ \lbrace x_{1},y_{1},\mu _{1},\nu _{1}\rbrace,\lbrace x_{2},y_{2},\mu _{2},\nu _{2}\rbrace,\ldots,\lbrace x_{l},y_{l},\mu _{l},\nu _{l}\rbrace \Bigl\} .\end{eqnarray*}
\subsubsection{Score value}
The score value for the training points is defined as:
\begin{align} \label{IF6} k_{i}=\begin{cases}
\mu _{i}, & \nu _{i}=0,\\ 0, & \mu _{i}\leq \nu _{i},\\ \frac{1-\nu _{i}}{2-\mu _{i}-\nu _{i}}, & \text{others}. \end{cases}  \end{align}

\section{Proposed work}
This section presents the proposed fuzzy membership assignment technique, and also discusses the linear and nonlinear formulations of the proposed algorithm.
\subsection{Proposed intuitionistic fuzzy assignment}
Since we are dealing with class imbalance, so we incorporate an imbalance ratio term in the existing score value (\ref{IF6}). This makes the score more effective in the class imbalance scenario.\\
The proposed fuzzy-based assignment to the slack factor is given by 
\begin{align} \label{IF7}
S_{i}=\begin{cases}k_i, & x \in \text{majority class,}\\ IR \times k_i, & x \in  \text{minority class}.\end{cases} 
\end{align} 
where $IR$ is defined as:
\begin{align} \label{IR}
   IR =\frac{\text {Total samples of majority class }}{\text {Total samples of minority class}}. 
\end{align}
In equation (\ref{IR}), the majority class is also referred as positive class, and majority class as negative class.

\subsection{Proposed projection based fuzzy membership assignment (PFMA)}
The existing fuzzy membership functions employed in various SVM variants exhibit certain limitations. One such drawback observed in the class center-based fuzzy membership is that these function assigns membership values solely based on the distance from the class center, disregarding whether the data point is closer to the proximal hyperplane of its own class or not. The proposed PFMA is based on the projection of the data point on the proximal hyperplane of its class.
Nasiri et al. \cite{nasiri2014energy} proposed an energy-based least square twin support vector machine LSTSVM (ELS-TSVM) that reduces the influence of noise and outliers by incorporating an energy component for each hyperplane. We get our proximal hyperplanes by using ELS-TSVM, as mentioned in equations (\ref{EL4}) and (\ref{EL5}). We use SMW  \cite{golub1996matrix}  formula to reduce computational cost and also incorporate $\delta>0$ in both equations to avoid ill-conditioning and singularity issues of the matrices in (\ref{EL4}) and (\ref{EL5}). Then we calculate the projection of the data point on its class hyperplane and denote it by $d_{hyp}$.
\begin{align}
 d_{hyp}(x)= \frac{\mid w_{i}^{T}x+b^{}_{i} \mid }{\| w_{i} \| }.
\end{align}

The proposed score value $h_i : X \rightarrow \mathbb{R} $ is defined as follows:
\begin{align} \label{f6}
h(x_{i})= \text{exp}{\left(- \frac{d_{hyp}(x_{i})-\min\limits_{x_{i} \in X}(d_{hyp}(x_{i}))}{\max\limits_{x_{i} \in X}(d_{hyp}(x_{i}))-\min\limits_{x_{i} \in X}(d_{hyp}(x_{i}))}\right)}.\end{align}
To deal with the imbalance issue in classification, we introduce an imbalance ratio term in the proposed fuzzy score value.
The training points are transformed into fuzzy numbers,
\begin{eqnarray*} T= \Bigl\{ \lbrace x_{1},y_{1},h _{1}\rbrace,\lbrace x_{2},y_{2},h_{2}\rbrace,\ldots,\lbrace x_{l},y_{l},h_{l}\rbrace \Bigl\} .\end{eqnarray*}
Hence, our assigned fuzzy-based weight is
$S_i : X \rightarrow \mathbb{R} $ 
\begin{align} \label{f7}
S_{i}=\begin{cases}h_i, & x \in \text{majority class,}\\ IR \times h_i, & x \in  \text{minority class}, \end{cases}  
\end{align}
where $IR$ is the same as defined in equation (\ref{IR}).\\
\textbf{Proposition $1$ :} $h(x)= \text{exp}{\left(- \frac{d_{hyp}(x)-\min\limits_{x \in X}(d_{hyp}(x))}{\max\limits_{x \in X}(d_{hyp}(x))-\min\limits_{x \in X}(d_{hyp}(x))}\right)}$ is a fuzzy function.\\
\textbf{Proof :}
$$ \min\limits_{x \in X}(d_{hyp}(x)) \leq d_{hyp}(x) \leq \max\limits_{x \in X}(d_{hyp}(x)).$$
Therefore 
$$0 \leq d_{hyp}(x)-\min\limits_{x \in X}(d_{hyp}(x)) \leq \max\limits_{x \in X}(d_{hyp}(x))-\min\limits_{x \in X}(d_{hyp}(x)).$$ 
$$ \implies 0 \leq \frac{d_{hyp}(x)-\min\limits_{x \in X}(d_{hyp}(x))}{\max\limits_{x \in X}(d_{hyp}(x))-\min\limits_{x \in X}(d_{hyp}(x))} \leq 1, $$

or equivalently 
$$0 \geq -\frac{d_{hyp}(x)-\min\limits_{x \in X}(d_{hyp}(x))}{\max\limits_{x \in X}(d_{hyp}(x))-\min\limits_{x \in X}(d_{hyp}(x))} \geq -1. $$
Since exponential is an increasing function, we have 
$$ e^{-1} \leq h_{i}(x)= \text{exp}{\left(- \frac{d_{hyp}(x)-\min\limits_{x \in X}(d_{hyp}(x))}{\max\limits_{x \in X}(d_{hyp}(x))-\min\limits_{x \in X}(d_{hyp}(x))}\right)} \leq 1.$$ 
Hence, $h(x)$ takes values only in the interval $[e^{-1}, 1]$. Since $[e^{-1}, 1] \subseteq [0, 1]$, so the function $h(x)$ is a fuzzy membership function.
\subsubsection{Properties of the proposed PFMA:}
\begin{figure}[H]
\centering
\begin{subfigure}{.5\textwidth}
  \centering
  \includegraphics[width=1.0\linewidth]{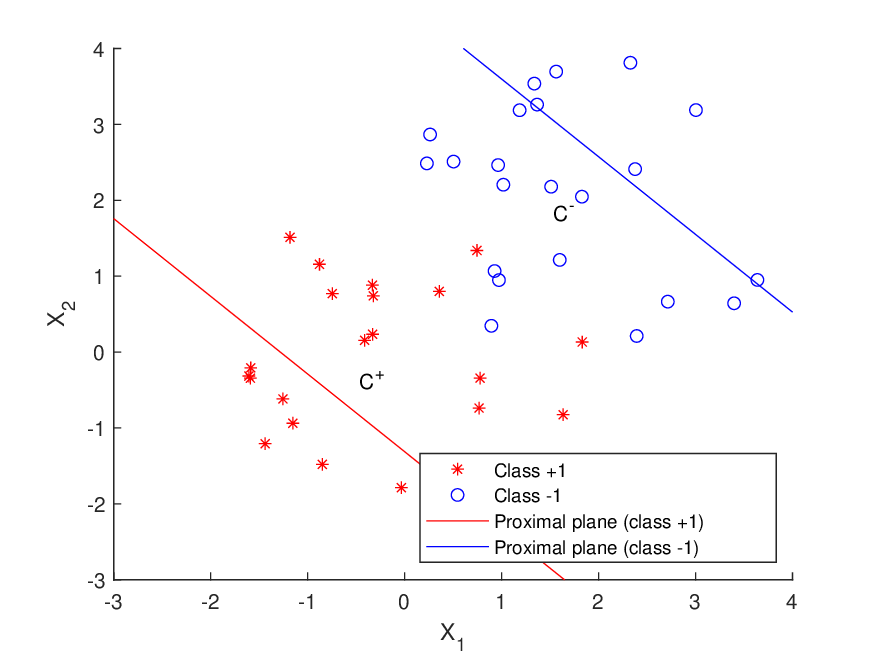}
   \caption{Plot of the proximal hyperplane of the binary artificial dataset.}  
  \label{PF2}
\end{subfigure}%
\begin{subfigure}{.5\textwidth}
  \centering
  \includegraphics[width=1.0\linewidth]{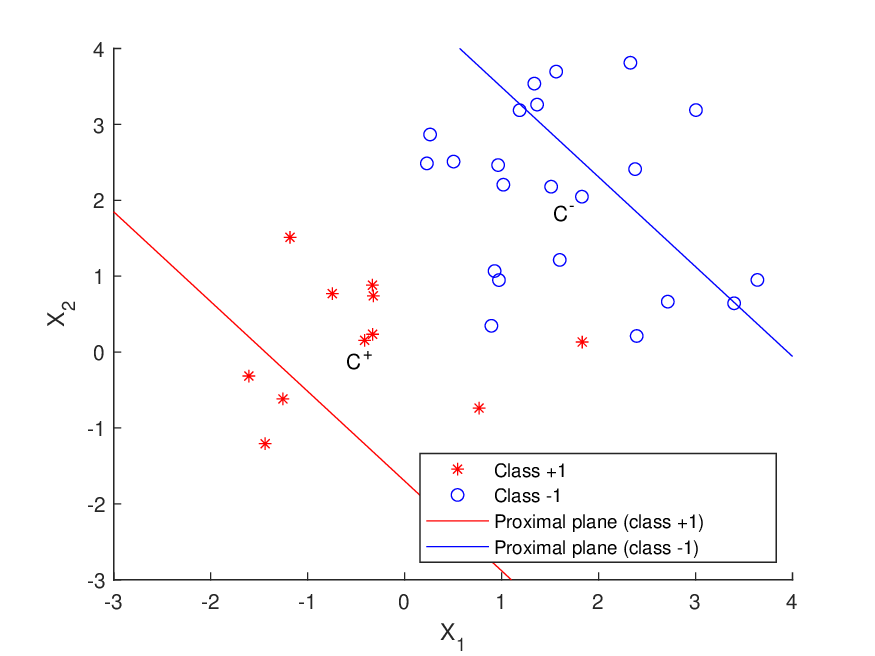}
    \caption{Plot of the shifted proximal hyperplane of the binary reduced artificial dataset.}  
  \label{PF5}
\end{subfigure}
\caption{Plot showing the shift in both the proximal hyperplanes with change in only one class. The hyperplanes are generated using the proposed PFMA approach.}
\label{fig: PF}
\end{figure}

\begin{figure}[H]
    \centering
    \includegraphics[width=4.0in,height=2.8in]{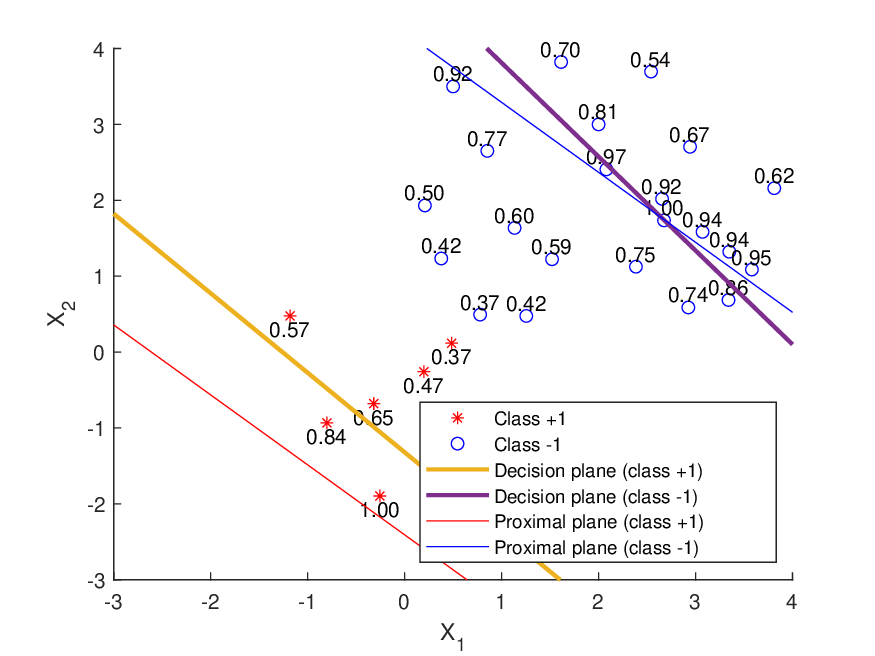}
    \caption{Plot showing membership values based on PFMA for an artificial dataset (IR = 3.67).}  
  \label{PF4}
\end{figure}

We discuss some properties of the proposed fuzzy membership function (\ref{f6}).
\begin{enumerate}

\item PFMA is based on proximal planes that are aligned to the distribution of data points, so the PFMA approach gives information about the data distribution for the fuzzy membership. This is in contrast to the centroid-based approach, where no information about data distribution is utilized for assigning the fuzzy membership.
    \item The data point farthest from the proximal hyperplane corresponds to the minimum membership value. 
    \item PFMA takes values in the range $[e^{-1},1]$.
    \item One can observe from Fig. (\ref{fig: PF}) that the proximal planes are influenced by the data points of the other class in the proposed PFMA.

\end{enumerate}
 Some cases for the proposed PFMA approach are discussed below (see Fig. \ref{PF4}).\\
\textbf{Case 1:} Higher membership values are obtained for data points nearer to the proximal plane of their own class.\\
\textbf{Case 2:} Lower membership values are obtained for data points that are farther from their own proximal plane. The membership value decreases as the data point moves farther from its own proximal plane.\\
\textbf{Case 3:} Data points far from the proximal plane of its own class or close to the opposite class hyperplane get very low membership values.


\subsubsection{IFMA vs proposed PFMA} 
\begin{figure}[H]
    \centering
    \includegraphics[width=4.0in,height=2.8in]{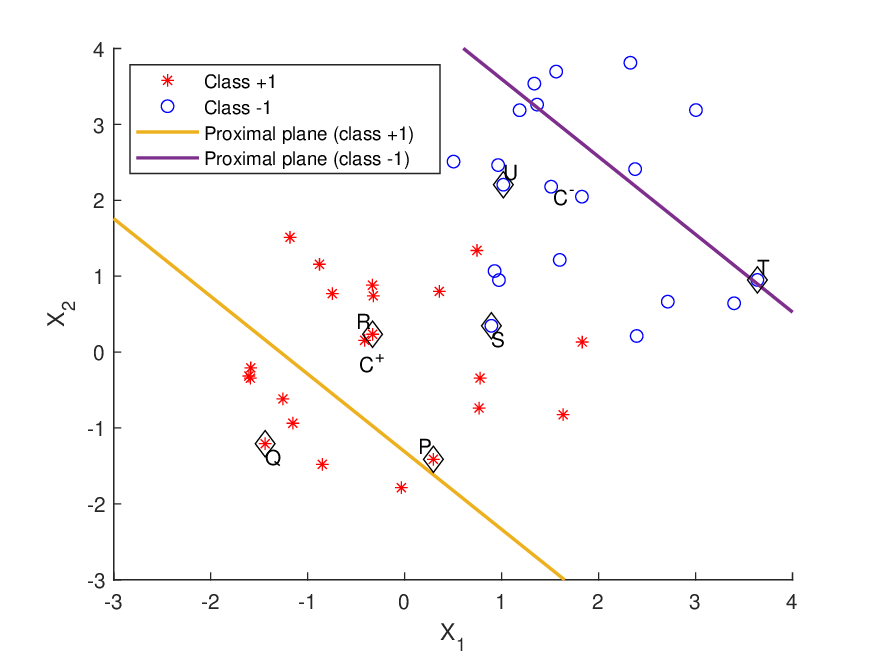}
    \caption{Fuzzy membership based on projection on proximal planes.}  
  \label{PF}
\end{figure}

    The IFMA approach assigns identical values to all the points having the same distance from the class centers. This technique awards the membership values based only on the distance from the centers. In contrast, the proposed PFMA distinguishes between the points based on their distance from the fixed proximal hyperplane. The points closest to the hyperplane get the highest membership values while the farthest point gets the least membership values. For instance, as shown in Fig. (\ref{PF}) points P and Q get nearly the same membership values in the IFMA approach but in the proposed PFMA approach point P has a higher membership value than point Q. 
   \par A noticeable difference between the two techniques is that it is possible in the new proposed PFMA for a point near the center to get a lower membership value. To illustrate, consider the points P, Q, and R in Fig. (\ref{PF}). In the IFMA approach, point R would get the higher membership among the three points P, Q, and R. On the other hand, in the proposed PFMA, point P would get the highest membership among the three, followed by R and then Q. This is due to the plane based membership calculation in the proposed PFMA, while in centroid-based membership functions the membership is based on distance from the class' centre (see Fig. \ref{fig: PF}).

\subsection{Proposed algorithms}
By including regularisation terms into the framework of ELS-TSVM, we propose IF-RELSTSVM and F-RELSTSVM by using IFMA (equation (\ref{IF7})) and proposed PFMA (equation (\ref{f6})), respectively. The linear and non-linear cases are discussed below:
\subsubsection{Linear case}
The optimization of the proposed algorithm is defined as:
\begin{align} \label{PROP1}&\underset{w_{1},b_{1},\xi _{2}}{\min }\,\frac{1}{2}\parallel Aw_{1}+e_{1}b_{1}\parallel ^{2}+ \frac{C_{1}}{2}(S_2\xi_{2})^{T}S_2\xi _{2} +\frac{1}{2}C_2(\parallel w_{1}\parallel ^{2}+ b_{1} ^{2})\nonumber \\ &\text{subject to}\ -(Bw_{1}+e_{2}b_{1})+\xi _{2} = E_{2},\, \end{align}
and 
\begin{align} &\underset{w_{2},b_{2},\xi _{1}}{\min }\,\frac{1}{2}\parallel Bw_{2}+e_{2}b_{2}\parallel ^{2}+ \frac{C_{3}}{2}(S_1\xi_{1})^{T}S_1\xi _{1} +\frac{1}{2}C_4(\parallel w_{2}\parallel ^{2}+ b_{2} ^{2})\nonumber \\&\text{subject to}\ (Aw_{2}+e_{1}b_{2})+\xi _{1} = E_{1},\, \end{align}
where $S_i$ is fuzzy-based weights to penalties, $C_1, C_2, C_3$ and $C_4$ are regulariztion parameters and $E_1$ , $E_2$ are energy parameters.\\
The Lagrangian for (\ref{PROP1}) is given by
\begin{align}\label{PROP4} L=\frac{1}{2}\parallel Aw_{1}+e_{1}b_{1}\parallel ^{2}+C_{1}\parallel S_2(Bw_{1}+e_{2}b_{1}+E_{2})\parallel ^{2}+\frac{1}{2}C_2(\parallel w_{1}\parallel ^{2}+ b_{1}^{2}). \end{align}
Setting the gradient of (\ref{PROP4}) with respect to $w_1$, $b_1$ and then equating it to zero gives:
\begin{align} \frac{\partial L}{\partial w_{1}}&=A^{T}(Aw_{1}+e_{1}b_{1})+C_{1} (S_2B)^TS_2(Bw_{1}+e_{2}b_{1}+E_{2})+C_2w_1=0 ,\\ \frac{\partial L}{\partial b_{1}}&=e_{1}^{T}(Aw_{1}+e_{1}b_{1})+C_{1} (S_2B)^TS_2(Bw_{1}+e_{2}b_{1}+E_{2})+C_2b_1=0 .\end{align} 
Combining the above both equations, we get
\begin{align}
\begin{bmatrix}
    A^{T} \\ e_{1}^{T} 
\end{bmatrix}\begin{bmatrix}
    A^{T}&e_{1}^{T} 
\end{bmatrix}\begin{bmatrix}
    w_{1} \\ b_{1}
\end{bmatrix}+ C_1 \begin{bmatrix}
   S_2 B^{T} \\ S_2 e_{2}^{T}
\end{bmatrix}   \begin{bmatrix} \begin{bmatrix} S_2 B\,\,S_2 e_{2} \end{bmatrix} \begin{bmatrix} w_{1} \\ b_{1} \end{bmatrix} + S_2 E_2\end{bmatrix}  +C_2 \begin{bmatrix} w_1 \\ b_{1} \end{bmatrix} =0.  \end{align} \\
Let
$H_1=\begin{bmatrix}A & e\end{bmatrix}$ and $G_2=\begin{bmatrix}S_2B & S_2e\end{bmatrix}$, then (\ref{PROP1}) can be reformulated as:

\begin{align} \begin{bmatrix}w_1 \\ b_1\end{bmatrix}=-\left(\frac{1}{C_1}H_{1}^{T}H_{1}+G_{2}^{T}G_{2}+\frac{C_{2}}{C_1}I\right)^{-1}G_{2}^{T}S_2E_2 . \end{align}

  Similarly, the solution for the other class is given by
\begin{align} \begin{bmatrix}w_2 \\ b_2\end{bmatrix}=\left(\frac{1}{C_{3}}J_{2}^{T}J_{2}+R_{1}^{T}R_{1}+\frac{C_{4}}{C_3}I\right)^{-1}R_{1}^{T} S_1 E_1, \end{align}
where
$R_1=\begin{bmatrix}S_1A & S_1e\end{bmatrix}$ and  $J_2=\begin{bmatrix}B & e\end{bmatrix}$. The linear IF-RELSTSVM training step is complete after acquiring two hyperplanes, $(w_1, b_1)$ and $(w_2, b_2)$. We obtain the label of a new data point $x$  using equation (\ref{DF}).

\subsection{Non-linear case}
In this section, we explore kernel-generated surfaces in place of planes and extend our results to nonlinear classifiers. Consider
$$
K\left(x^T, C^T\right) w_{(1)}+b_{(1)}=0 \text{ and } K\left(x^T, C^T\right) w_{(2)}+b_{(2)}=0,
$$
where $C^T=\begin{bmatrix}A&; B\end{bmatrix}^T$,
and $K$ is the kernel function. The optimization problems for the non-linear case are written as:
\begin{align} \label{PROP1x} &\underset{w_{1},b_{1},\xi _{2}}{\min }\,\frac{1}{2}\parallel K(A,C^T)w_{1}+e_{1}b_{1}\parallel ^{2}+C_{1}(S_2\xi_{2})^{T}S_2\xi _{2} +\frac{1}{2}C_2(\parallel w_{1}\parallel ^{2}+ b_{1} ^{2})\nonumber \\ &\text{subject to}\ -(K(B,C^T)w_{1}+e_{2}b_{1})+\xi _{2} = E_{2},\, \end{align}
and 
\begin{align} \label{PROP2x} &\underset{w_{2},b_{2},\xi _{1}}{\min }\,\frac{1}{2}\parallel K(B,C^T)w_{2}+e_{2}b_{2}\parallel ^{2}+C_{3}(S_1\xi_{1})^{T}S_1\xi _{1} +\frac{1}{2}C_4(\parallel w_{2}\parallel ^{2}+ b_{2} ^{2})\nonumber \\&\text{subject to}\ (K(A,C^T)w_{2}+e_{1}b_{2})+\xi _{1} = E_{1},\, \end{align} 
where the sizes of kernel matrices $K(A,C^T)$ and $K(B,C^T)$ are $p \times m$ and $q \times m$ respectively, $ m = p + q$.\\
The Lagrangian for (\ref{PROP1x}) is given by
\begin{align} \label{PROP3x} L(w_{1},b_{1},\xi _{2},\alpha,\beta)=\frac{1}{2}\parallel K(A,C^T)w_{1}+e_{1}b_{1}\parallel ^{2}+C_{1}(S_2\xi_{2})^{T}S_2\xi _{2}+\frac{1}{2}C_2(\parallel w_{1}\parallel ^{2}+ b_{1} ^{2}). \end{align}
Putting the values of $\xi_2$ in lagrangian (\ref{PROP3x}), we get
\begin{align} \label{PROP4x} L=&\frac{1}{2}\parallel K(A,C^T)w_{1}+e_{1}b_{1}\parallel ^{2}+ \\ \nonumber &C_{1}\parallel S_2(K(B,C^T)w_{1}+e_{2}b_{1}+E_{2})\parallel ^{2}+\frac{1}{2}C_2(\parallel w_{1}\parallel ^{2}+ b_{1} ^{2}). \end{align}
Setting the gradient of (\ref{PROP4x}) with respect to $w_1$ and $b_1$ to zero gives:
\begin{align} \frac{\partial L}{\partial w_{1}}=&K(A,C^T)^{T}(K(A,C^T)w_{1}+e_{1}b_{1})+ \\ \nonumber  &C_{1} (S_2 K(B,C^T))^TS_2(K(B,C^T)w_{1}+e_{2}b_{1}+E_{2})+C_2w_1=0 ,\\ \frac{\partial L}{\partial b_{1}}=&e_{1}^{T}(K(A,C^T)w_{1}+e_{1}b_{1})+ \\ \nonumber &C_{1} (S_2 K(B,C^T))^TS_2( K(B,C^T) w_{1}+e_{2}b_{1}+E_{2})+C_2b_1=0. \end{align} 
Combining the above two equations, we get
\begin{align} &\begin{bmatrix}
    K(A,C^T)^{T} \\ e_{1}^{T}
\end{bmatrix}\begin{bmatrix}
    K(A,C^T)&e_{1}
\end{bmatrix}\begin{bmatrix}
    w_{1} \\ b_{1}
\end{bmatrix}+ 
\\ \nonumber & C_1  \begin{bmatrix}
    S_2 K(B,C^T)^{T} \\ S_2 e_{2}^{T}
\end{bmatrix} \begin{bmatrix}\begin{bmatrix}
    S_2 K(B,C^T)& S_2 e_{2}
\end{bmatrix}\begin{bmatrix}
    w_{1} \\ b_{1}
\end{bmatrix} S_2 E_2 
\end{bmatrix} + C_2 \begin{bmatrix}
    w_1 \\ b_{1}
\end{bmatrix}=0.  \end{align} 
Let
$H_1=\begin{bmatrix}K(A,C^T)&e_{1}\end{bmatrix}$ and $G_2=\begin{bmatrix}S_2 K(B,C^T)& S_2 e_{2}\end{bmatrix}$, then (\ref{PROP1x}) can be reformulated as 
 \begin{align} \begin{bmatrix}w_1 \\ b_1\end{bmatrix}=-\left(\frac{1}{C_1}H_{1}^{T}H_{1}+G_{2}^{T}G_{2}+\frac{C_{2}}{C_1}I\right)^{-1}G_{2}^{T}S_2E_2 . \end{align}
In a similar way, solution for other class is given by
\begin{align} \begin{bmatrix}w_2 \\ b_2\end{bmatrix}=\left(\frac{1}{C_{3}}J_{2}^{T}J_{2}+R_{1}^{T}R_{1}+\frac{C_{4}}{C_3}I\right)^{-1}R_{1}^{T} S_1 E_1 . \end{align}
where
$R_1=\begin{bmatrix}
    S_1 K(A,C^T) & S_1e_1 \end{bmatrix}$ and $ J_2=\begin{bmatrix}K(B,C^T)& e_{2}\end{bmatrix}$.
    
Note: $(\frac{1}{C_{3}}J_{2}^{T}J_{2}+R_{1}^{T}R_{1}+\frac{C_4}{C_3} I)$ and $(\frac{1}{C_1}H_{1}^{T}H_{1}+G_{2}^{T}G_{2}+\frac{C_2}{C_1} I)$ are positive definite, which makes our model stable and robust than LSTSVM and ELS-TSVM. Moreover, the solution of nonlinear F-RELSTSVM and IF-RELSTSVM requires the two matrix inverse of size $(m + 1) \times (m+1)$. As a result, the SMW formula is employed to approximate and decrease the above equation's computation costs.\\\\
\textbf{Case 1:} $p < q$
\begin{align}
& {\left[\begin{array}{ll}w_1 \\ b_1\end{array}\right]=-\left(Y-Y H_{1}^t\left(C_1 I+H_{1} Y H_{1}^t\right)^{-1} H_{1} Y\right) V^t E_2},\end{align}
$\text { where }$
 $Y=\frac{C_1}{C_2}\left(I-G_{2}^t\left(\frac{C_2}{C_1}I+G_{2} G_{2}^t\right)^{-1} G_{2}\right)$.
\begin{align}
& {\left[\begin{array}{ll}w_2 \\ b_2\end{array}\right]=C_2\left(Y_1-Y_1 J_{2}^t\left(\frac{I}{C_2}+J_{2} Y_1 J_{2}^t\right)^{-1} J_{2} Y_1\right) J_{2}^t E_1},
\end{align}
$\text { where } $
 $Y_1=\frac{C_3}{C_4}\left(I-R_{1}^t\left(\frac{C_4}{C_3} I+R_{1} R_{1}^t\right)^{-1} R_{1}\right)$.\\
\textbf{Case 2:} $q < p$
 \begin{align} & {\left[\begin{array}{ll}w_1 \\ b_1\end{array}\right]=-C_1\left(Z-Z G_{2}^t\left(\frac{I_2}{C_1}+G_{2} Z^t G_{2}\right)^{-1} G_{2} Z\right) G_{2}^t E_2}, 
 \end{align}
 where
  $
  Z=\frac{C_1}{C_2}\left(I-H_{1}^t\left(\frac{C_2}{C_1} I+H_{1} H_{1}^t\right)^{-1} H_{1}\right).
  $
\begin{align}
 {\left[\begin{array}{ll}w_2 \\ b_2\end{array}\right]=\left(Z_1-Z_1 R_{1}^t\left(C_2 I+R_{1} Z_1 R_{1}^t\right)^{-1} R_{1} Z_1\right) R_{1}^t E_{1}},\end{align}
 \text{where} 
$
Z_1=\frac{C_3}{C_4}\left(I-J_{2}^t\left(\frac{C_4}{C_3} I+J_{2} J_{2}^t\right)^{-1} J_{2}\right).
$

The label of an unknown data point $x$ is assigned to class $ (i = +1, -1)$ based on the decision function below.
\begin{align} \text{Class label}(x)=\begin{cases}+1, &\frac{\mid K(x_i,C^T)w_{1}+b_{1}\mid }{\mid K(x_i,C^T)w_{2}+b_{2} \mid } \leq 1,\\ -1,&\frac{\mid K(x_i,C^T)w_{1}+b_{1}\mid }{\mid K(x_i,C^T)w_{2}+b_{2} \mid } > 1, \end{cases}  \end{align}

\textbf{Proposition $2$ :}
For any given $C_1$ and $C_2>0$, $\left\{\frac{1}{C1}H_1^T H_1+\frac{C_2}{C_1} I+ G_2^T G_2\right\}$ is a invertible matrix.\\
\textbf{Proof :} Let $x$ be a non-zero column vector of $(m+1) \times 1$ order.
Now,
$$
\begin{array}{r}
x^T\left(H_1^T H_1\right) x=(H_1 x)^T(H_1 x),\\
=\|H_1 x\|^2 \geq 0.
\end{array}
$$
Since for any $C_1>0,$ $\frac{1}{C_1}H_1^T H_1$ and $G_2^T G_2$ are positive semi-definite matrices, so $\left\{\frac{1}{C1}H_1^T H_1+ G_2^T G_2\right\}$ is a positive semi-definite matrix.
Now, as $I$ (identity) is a positive definite matrix, therefore, for all $C_1$ and $C_2>0$, $\left\{\frac{1}{C1}H_1^T H_1+\frac{C_2}{C_1} I+ G_2^T G_2\right\}$ is a positive definite matrix, and hence invertible.

\subsection{Computational complexity}
Here, we describe the computation complexity of the proposed algorithm. The proposed algorithm maintains the same size for the invertible matrices as in LSTSVM, thereby avoiding additional computational overhead when solving the optimization problem. In the LSTSVM formulation, two matrix inverses are calculated of size $(m + 1) \times (m + 1)$, where $m$ represents the sum of the number of data points in the positive and negative classes (denoted as $p$ and $q$, respectively). To streamline the computation of inverses, the Sherman-Morrison-Woodbury (SMW) formula \cite{golub1996matrix} is employed. This formula allows us to solve three inverses of smaller sizes. 
\par The computation required for the proposed algorithms involves the fuzzy membership degree calculation. However, our IFMA has a time complexity of $O(m)$, as they calculate fuzzy membership for all data points using measures like distance from the centroid. 
\par The proposed PFMA involves matrix inverse, in the case where $p < q$, the nonlinear PFMA requires two matrix inverses of size $(p \times p)$ and one matrix inverse of size $(q \times q)$. Conversely, when $q < p$, the algorithm necessitates two matrix inverses of size $(q \times q)$ and one matrix inverse of size $(p \times p)$.

\section{Numerical experiments}
In this section, we perform numerical experiments to compare the proposed F-RELSTSVM and IF-RELSTSVM methods with the baseline models and also perform statistical analysis to demonstrate the significance of the proposed models.
\subsection{Parameter selection} 
We demonstrate the effectiveness of the proposed fuzzy membership assignment by comparing the proposed F-RELSTSVM and IF-RELSTSVM with other baseline algorithms such as LSTSVM \cite{ARUNKUMAR20097535}, ELS-TSVM \cite{nasiri2014energy}, RELS-TSVM \cite{tanveer2016robust}, and IFLSTSVM \cite{laxmi2022intuitionistic} on different synthetic and real world imbalanced datasets. 
To determine the best parameters, a five-fold cross-validation technique, along with grid search, is employed for all the algorithms. The Gaussian kernel is given by $K({x_i}, {x_k}) = \exp\left(-\frac{\|{x_i} - {x_k}\|^2}{2\sigma}\right)$, where ${x_i}, {x_k} \in \mathbb{R}^m$ are the input vectors and $\sigma$ is the kernel parameter. To compare the algorithms, the AUC \cite{huang2005using} is calculated as, $$\text{AUC} = \frac{1+T_P-F_P}{2}.$$
Here, $T_P$ represents the true positive rate and $F_P$ represents the false positive rate.
The value of Gaussian kernel parameter $\sigma$
for all the algorithms is selected from the set $\lbrace{2^i
|i = -5, -4, ..., 5}\rbrace$. The
value of the parameters $C_i, i=\{1,2,3,4\}$ and $E_i, i=\{1,2\}$ are chosen from the sets $\lbrace{10^i
|i = -5, -4, ..., 5}\rbrace$ and $\lbrace 0.6, 0.7, 0.8, 0.9, 1 \rbrace$
respectively for both linear and nonlinear kernels. 
In order to reduce the training cost, we set $C_1 = C_2$ for LSTSVM, ELS-TSVM, and $C_1 = C_2, C_3 = C_4$ for IFLSTSVM, RELS-TSVM, F-RELSTSVM, and IF-RELSTSVM. In this study, the AUC values for all the algorithms are expressed as percentages.

\subsection{Real world datasets}
To assess the performance of the proposed IF-RELSTSVM and F-RELSTSVM algorithms, we compare proposed algorithms to LSTSVM, ELS-TSVM, IFLSTSVM, and RELS-TSVM on $29$ datasets from the KEEL repository \cite{derrac2015keel, napierala2010learning}. We can observe from Table (\ref{table:2}) and (\ref{table:1}) that for linear kernel, the proposed IF-RELSTSVM and F-RELSTSVM outperform  all other baseline algorithms with higher average AUC, $i.e.$ $81.94$ and $82.21$, respectively. The average ranks of all the algorithms using AUC were calculated and presented in Table (\ref{table:3}).
\subsection{Statistical analysis on real world datasets}
In this sub-section, we perform a statistical analysis of the compared models to demonstrate the significance of the proposed IF-RELSTSVM and F-RELSTSVM models. The statistical tests are performed using the Friedman and Nemenyi post-hoc test \cite{demvsar2006statistical}.

\textbf{Friedman test}: We statistically assess the models using the Friedman test. Under the null hypothesis, we assume that all the algorithms are similar, $i.e.$, the average rank of all algorithms is equal. The Friedman test follows the chi-squared distribution ($\chi^2_{F}$) with ($k-1$) degrees of freedom ($k$ is the number of models), and $N$ datasets. The chi-squared ($\chi^2_{F}$) distribution is calculated using Table (\ref{table:3}) as,
\begin{align}
\chi_F^2=\frac{12N}{k(k+1)}\left[\sum_{j=1}^k R_j^2-\frac{k(k+1)^2}{4}\right],
\end{align}
 \begin{align}
\chi_F^2=&\frac{12 \times 29}{6(6+1)}\left[4.47^2+3.57^2+3.45^2+3.47^2+3.05^2+3^2-\frac{6(7)^2}{4}\right]\\=&11.52, \nonumber\\
F_F=&\frac{(N-1) \chi_F^2}{N(k-1)-\chi_F^2}=\frac{(29-1) \times 11.52 }{29(6-1)-11.52}=2.42 .
\end{align}
 where the $F_F$ value is as per the $F$-distribution with degrees of freedom $(k - 1)$ and $(k - 1)(N - 1),$ i.e.,$(5, 140)$, as we are considering the inclusion of $6$ algorithms and $29$ datasets with linear kernel. For a significance level of $\alpha$ = $0.05$, the critical value of $F (5, 140)$ amounts to $2.28$. Since the obtained $F_F$ value surpasses the critical value, we reject the null hypothesis. \\
Also, to examine the significant difference between the algorithms we employ the Nemenyi post-hoc test.

\begin{figure}
  \includegraphics[width=5.0in,height=2.5in]{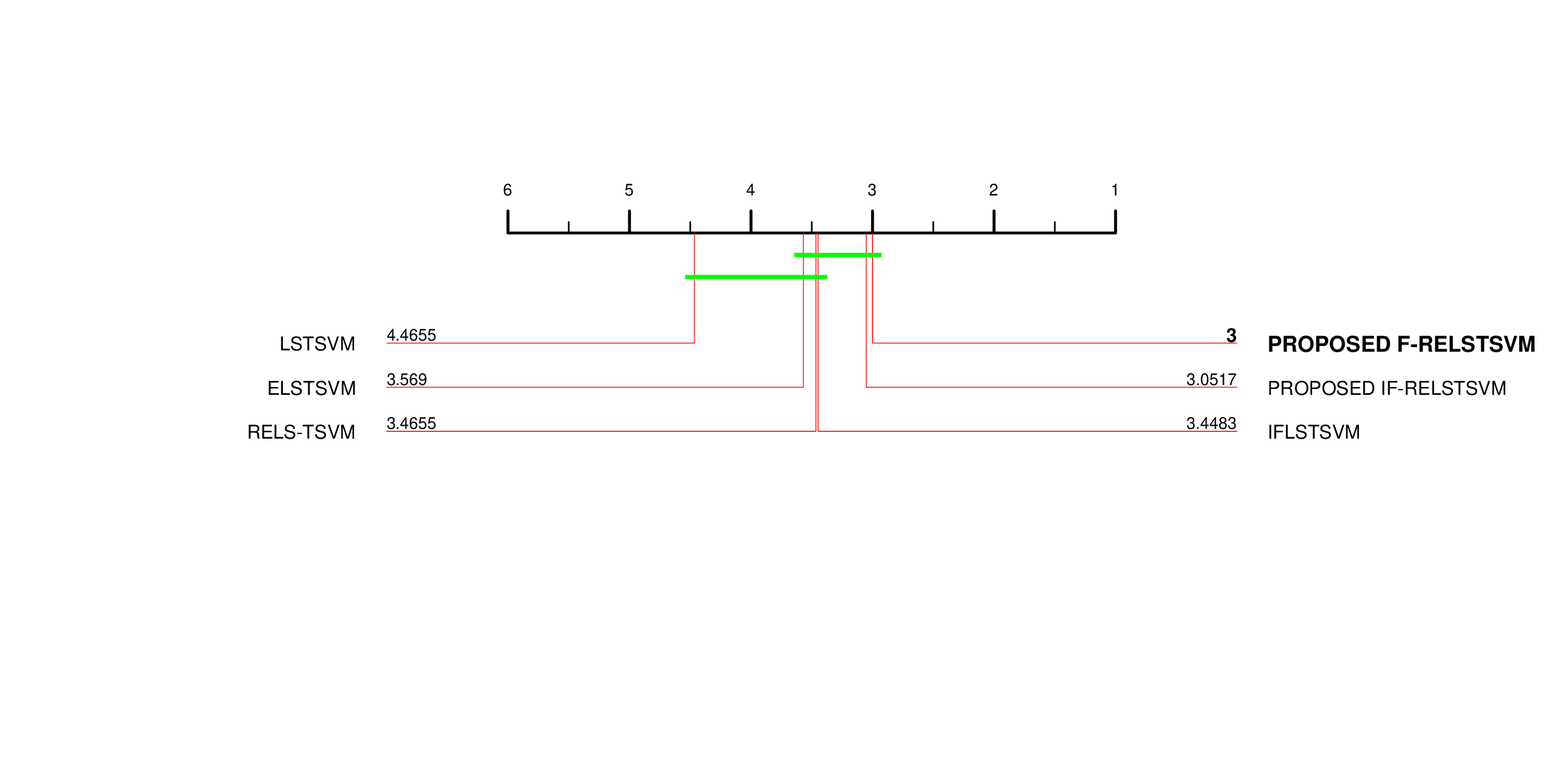}
    \caption{Critical difference diagram with linear kernel on real world datasets.}
     \label{fig:CD2}
\end{figure}

\textbf{Nemenyi post-hoc test}:
To perform pairwise comparisons, we utilize the Nemenyi post-hoc test. At a significance level of $p = 0.10$, $CD$ is calculated to be $1.27$.
$$CD=2.589 \times \sqrt{\frac{6 \times 7 }{ 6 \times 29}}=1.27.$$
As evident from Fig. (\ref{fig:CD2}), the proposed F-RELSTSVM algorithm has a significant difference ($1.46$) with LSTSVM, while no significant difference with the other baseline algorithms for the linear kernel. Also, it can be seen that the proposed IF-RELSTSVM is the second best in terms of average rank among these algorithms.
 \\
To further compare the performance of six algorithms, we statistically analyze the AUC using a Gaussian kernel. The average ranks of all the algorithms based on their AUC are computed and presented in Table \ref{table:4}. The Friedman statistic will be employed under the assumption of the null hypothesis.
\begin{align}
\chi_F^2=&\frac{12 \times 29}{6(6+1)}\left[3.69^2+4.10^2+4.09^2+3.39^2+3.39^2+2.33^2-\frac{6(7)^2}{4}\right],\\=&17.72 \nonumber \\
F_F=&\frac{(N-1) \chi_F^2}{N(k-1)-\chi_F^2}=\frac{(29-1) \times 17.72 }{29(6-1)-17.72}=3.90,
\end{align}

\begin{figure}
  \includegraphics[width=5.0in,height=2.5in]{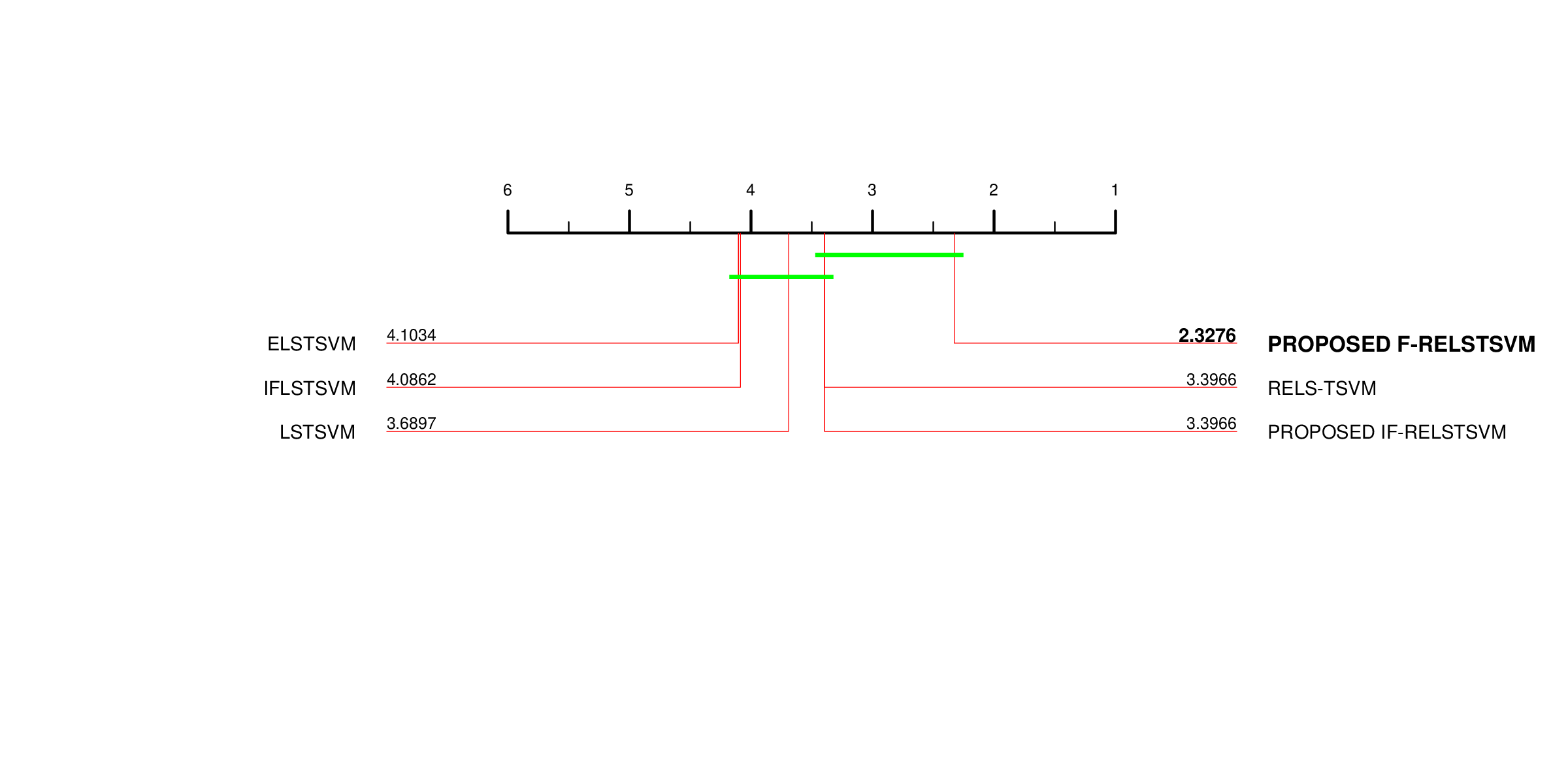}
    \caption{Critical difference diagram with Gaussian kernel on real world datasets.}
     \label{fig:CD1}
\end{figure}

Again, considering the inclusion of 6 algorithms and $29$ datasets, the computed $F_F$ value adheres to the F-distribution with degrees of freedom $(k - 1)$ and $(k - 1)(N - 1) = (5, 140)$. For a significance level of $\alpha$= $0.05$, the critical value for $F (5, 140)$ is $2.28$, and the $CD$ is $1.27$. The obtained $F_F$ value surpasses the critical value, we can reject the null hypothesis. Since the difference between the rank of proposed F-RELSTSVM to LSTSVM, ELS-TSVM, and IFLSTSVM is greater than $1.27, (3.69-2.33)=1.36,(4.10-2.33)= 1.77, (4.09-2.33)=	1.76$, respectively,
 hence the proposed F-RELSTSVM has a significant difference from the other baseline algorithms. Also, the proposed IF-RELSTSVM has lower average rank than LSTSVM, IFLSTSVM, amd ELSTSVM algorithms. Similar to the linear case, F-RELSTSVM and IF-RELSTSVM have lower ranks than all the other algorithms, as shown in Fig. (\ref{fig:CD1}). 

\begin{table}[]
\centering
{\caption{Performance comparison on real world datasets with linear kernel.}
 \label{table:2}
 \resizebox{1.1\textwidth}{!}{
\begin{tabular}{lcccccc}
\hline
DATASETS &
  LSTSVM \cite{ARUNKUMAR20097535}&
  ELS-TSVM \cite{nasiri2014energy} &
  IFLSTSVM \cite{laxmi2022intuitionistic} &
  RELS-TSVM \cite{tanveer2016robust} &
  \begin{tabular}[c]{@{}l@{}}Proposed\\ IF-RELSTSVM\end{tabular} &
  \begin{tabular}[c]{@{}l@{}}Proposed\\ F-RELSTSVM\end{tabular} \\

 &
  AUC(\%), Time(s) &
  AUC(\%), Time(s) &
  AUC(\%), Time(s) &
  AUC(\%), Time(s) &
  AUC(\%), Time(s) &
  AUC(\%), Time(s) \\
  \hline
Abalone9-18 &
  84.82$,$ 0.00699 &
  83.16$,$ 0.00221 &
  84.82$,$ 0.03212 &
  79.31$,$ 0.00722 &
  84.58$,$ 0.03643 &
  \textbf{85.31}$,$ 0.02254 \\
Brwisconsin &
  96.2$,$ 0.00432 &
  92.72$,$ 0.00039 &
  96.25$,$ 0.02154 &
  92.72$,$ 0.00251 &
  94.67$,$ 0.0245 &
  \textbf{97.43}$,$ 0.0139 \\
Bupa or liver-disorders &
  62.64$,$ 0.00152 &
  63.36$,$ 0.00039 &
  66.03$,$ 0.00549 &
  67.43$,$ 0.0013 &
  \textbf{69.99}$,$ 0.00884 &
  60.02$,$ 0.00511 \\
Crossplane150 &
  \textbf{100}$,$ 0.00243 &
  \textbf{100}$,$ 0.00009 &
  \textbf{100}$,$ 0.0028 &
  \textbf{100}$,$ 0.00025 &
  \textbf{100}$,$ 0.00533 &
  \textbf{100}$,$ 0.00627 \\
Ecoli\_0\_1\_4\_6-vs-5 &
  93.21$,$ 0.00104 &
  85.19$,$ 0.0001 &
  95.06$,$ 0.00303 &
  85.19$,$ 0.00089 &
  \textbf{96.91$,$ 0.00524} &
  96.3$,$ 0.00201 \\
Ecoli\_0\_1\_4\_7-vs-5\_6 &
  74.44$,$ 0.00072 &
  80$,$ 0.00009 &
  85.56$,$ 0.00417 &
  80$,$ 0.00105 &
  86.11$,$ 0.00414 &
  \textbf{86.67}$,$ 0.00218 \\
Ecoli\_0\_1-vs-2\_3\_5 &
  59.09$,$ 0.0005 &
  \textbf{86.07}$,$ 0.00014 &
  84.75$,$ 0.00252 &
  81.82$,$ 0.00051 &
  84.75$,$ 0.00274 &
  \textbf{86.07}$,$ 0.00157 \\
Ecoli\_0\_1-vs-5 &
  87.5$,$ 0.00049 &
  86.72$,$ 0.00007 &
  92.97$,$ 0.003 &
  \textbf{93.75}$,$ 0.00046 &
  \textbf{93.75}$,$ 0.00293 &
  82.03$,$ 0.00142 \\
Ecoli\_0\_2\_6\_7-vs-3\_5 &
  64.29$,$ 0.00075 &
  68.93$,$ 0.00008 &
  \textbf{71.07}$,$ 0.00233 &
  68.93$,$ 0.00117 &
  69.76$,$ 0.00293 &
  68.93$,$ 0.00152 \\
Ecoli\_0\_3\_4\_7-vs-5\_6 &
  84.85$,$ 0.0007 &
  84.85$,$ 0.00007 &
  84.85$,$ 0.00293 &
  80.3$,$ 0.00074 &
  \textbf{85.61}$,$ 0.00276 &
  \textbf{85.61}$,$ 0.00159 \\
Ecoli01vs5 &
  94.44$,$ 0.00071 &
  \textbf{99.21}$,$ 0.00007 &
  \textbf{99.21}$,$ 0.00319 &
  \textbf{99.21}$,$ 0.00068 &
  98.41$,$ 0.00343 &
  98.41$,$ 0.00139 \\
Ecoli4 &
  \textbf{97.31}$,$ 0.00068 &
  78.57$,$ 0.00007 &
  \textbf{97.31}$,$ 0.00442 &
  94.09$,$ 0.00107 &
  \textbf{97.31}$,$ 0.00768 &
  96.77$,$ 0.00212 \\
Glass\_0\_1\_4\_6-vs-2 &
  41.23$,$ 0.00041 &
  \textbf{77.85}$,$ 0.00007 &
  47.59$,$ 0.0039 &
  \textbf{77.85}$,$ 0.00067 &
  42.11$,$ 0.00288 &
  76.97$,$ 0.00133 \\
Glass\_0\_1\_6-vs-5 &
  92.45$,$ 0.0006 &
  98.11$,$ 0.00007 &
  \textbf{100}$,$ 0.00315 &
  \textbf{100}$,$ 0.00055 &
  96.23$,$ 0.00331 &
  89.62$,$ 0.00152 \\
Glass\_0\_6-vs-5 &
  90$,$ 0.00024 &
  95$,$ 0.00007&
  95$,$ 0.00209 &
  95$,$ 0.00041 &
  \textbf{100}$,$ 0.00164 &
  95$,$ 0.00069 \\
Glass2 &
  58.05$,$ 0.00066 &
  \textbf{81.32}$,$ 0.00008 &
  69.83$,$ 0.0025 &
  \textbf{81.32}$,$ 0.00071 &
  77.01$,$ 0.00338 &
  74.71$,$ 0.00143 \\
Glass4 &
  55.36$,$ 0.00065 &
  \textbf{83.93}$,$ 0.00007 &
  58.93$,$ 0.00264 &
  77.68$,$ 0.00059 &
  \textbf{83.93}$,$ 0.0044 &
  65.18$,$ 0.00154 \\
Haber &
  65.15$,$ 0.00088 &
  58.66$,$ 0.00006 &
  \textbf{73.82}$,$ 0.00241 &
  68.8$,$ 0.00059 &
  67.33$,$ 0.004 &
  73.08$,$ 0.00191 \\
Monk1 &
  \textbf{50}$,$ 0.00189 &
  \textbf{50}$,$ 0.00012 &
  44.73$,$ 0.00573 &
  47.25$,$ 0.00107 &
  48.38$,$ 0.00843 &
  \textbf{50}$,$ 0.0038 \\
Monk2 &
  50$,$ 0.00191 &
  63.89$,$ 0.00018 &
  50$,$ 0.00672 &
  64.68$,$ 0.00115 &
  66.01$,$ 0.009 &
  \textbf{73.28}$,$ 0.00429 \\
New-thyroid1 &
  98.21$,$ 0.00043 &
  98.21$,$ 0.00006 &
  98.21$,$ 0.00178 &
  \textbf{99.11}$,$ 0.00044 &
  \textbf{99.11}$,$ 0.0026 &
  \textbf{99.11}$,$ 0.00135 \\
Shuttle\_6-vs-2\_3 &
  \textbf{100}$,$ 0.00057 &
  \textbf{100}$,$ 0.00011 &
  \textbf{100}$,$ 0.00218 &
  \textbf{100}$,$ 0.00053 &
  99.25$,$ 0.00259 &
  99.25$,$ 0.0014 \\
Shuttle\_c0-vs-c4 &
  \textbf{100}$,$ 0.02903 &
  \textbf{100}$,$ 0.00027 &
  \textbf{100}$,$ 0.05638 &
  \textbf{100}$,$ 0.0308 &
  \textbf{100}$,$ 0.10022 &
  \textbf{100}$,$ 0.07682 \\
Transfusion &
  57.48$,$ 0.0041 &
  \textbf{64.15$,$ 0.00016} &
  55.54$,$ 0.00868 &
  62.67$,$ 0.00291 &
  60.56$,$ 0.01304 &
  59.04$,$ 0.00894 \\
Vehicle 1 &
  77.34$,$ 0.00465 &
  79.44$,$ 0.00024 &
  77.6$,$ 0.01558 &
  79.7$,$ 0.00362 &
  74.14$,$ 0.02129 &
  \textbf{80.76$,$ 0.01202} \\
Yeast\_0\_2\_5\_6-vs-3\_7\_8\_9 &
  71.86$,$ 0.00671 &
  76.3$,$ 0.00024 &
  75.19$,$ 0.02493 &
  74.82$,$ 0.00671 &
  \textbf{78.16}$,$ 0.03683 &
  76.49$,$ 0.02153 \\
Yeast\_0\_3\_5\_9-vs-7\_8 &
  63.19$,$ 0.00133 &
  \textbf{63.94}$,$ 0.00016 &
  62.08$,$ 0.00739 &
  \textbf{63.94}$,$ 0.00134 &
  62.82$,$ 0.0094 &
  62.82$,$ 0.00422 \\
Yeast1vs7 &
  68.33$,$ 0.00162 &
  \textbf{68.15}$,$ 0.00018 &
  75.87$,$ 0.01066 &
  68.15$,$ 0.00154 &
  66.16$,$ 0.0082 &
  71.68$,$ 0.00381 \\
Yeast3 &
  92.66$,$ 0.01549 &
  88.05$,$ 0.00026 &
  92.54$,$ 0.0497 &
  89.49$,$ 0.01651 &
  93.24$,$ 0.06955 &
  \textbf{93.49}$,$ 0.04338 \\
  \hline
Average AUC(\%), Average time(s) &
  76.9$,$ 0.00317 &
  81.23$,$ 0.00021 &
  80.51$,$ 0.01014 &
  81.83$,$ 0.00303 &
  \textbf{81.94$,$ 0.01406} &
  \textbf{82.21}$,$ 0.00868 \\
  \hline
\end{tabular}}}
\end{table}

\begin{table}[]
\centering{
\caption{Comparison on ranks based on AUC for real world datasets with linear kernel.}
 \label{table:3}
\resizebox{1.1\textwidth}{!}{
\begin{tabular}{lcccccc}
\hline
DATASETS &
  LSTSVM \cite{ARUNKUMAR20097535}&
  ELS-TSVM \cite{nasiri2014energy} &
  IFLSTSVM \cite{laxmi2022intuitionistic} &
  RELS-TSVM \cite{tanveer2016robust} &
  \begin{tabular}[c]{@{}l@{}}Proposed\\ IF-RELSTSVM\end{tabular} &
  \begin{tabular}[c]{@{}l@{}}Proposed\\ F-RELSTSVM\end{tabular} \\
  \hline
Abalone9-18                & 2.5      & 5        & 2.5      & 6        & 4        & 1   \\
Brwisconsin                & 3        & 5.5      & 2        & 5.5      & 4        & 1   \\
Bupa or liver-disorders    & 5        & 4        & 3        & 2        & 1        & 6   \\
Crossplane150              & 3.5      & 3.5      & 3.5      & 3.5      & 3.5      & 3.5 \\
Ecoli\_0\_1\_4\_6-vs-5       & 4        & 5.5      & 3        & 5.5      & 1        & 2   \\
Ecoli\_0\_1\_4\_7-vs-5\_6     & 6        & 4.5      & 3        & 4.5      & 2        & 1   \\
Ecoli\_0\_1-vs-2\_3\_5       & 6        & 1.5      & 3.5      & 5        & 3.5      & 1.5 \\
Ecoli\_0\_1-vs-5           & 4        & 5        & 3        & 1.5      & 1.5      & 6   \\
Ecoli\_0\_2\_6\_7-vs-3\_5     & 6        & 4        & 1        & 4        & 2        & 4   \\
Ecoli\_0\_3\_4\_7-vs-5\_6     & 4        & 4        & 4        & 6        & 1.5      & 1.5 \\
Ecoli01vs5                 & 6        & 2        & 2        & 2        & 4.5      & 4.5 \\
Ecoli4                     & 2        & 6        & 2        & 5        & 2        & 4   \\
Glass\_0\_1\_4\_6-vs-2       & 6        & 1.5      & 4        & 1.5      & 5        & 3   \\
Glass\_0\_1\_6-vs-5         & 5        & 3        & 1.5      & 1.5      & 4        & 6   \\
Glass\_0\_6-vs-5           & 6        & 3.5      & 3.5      & 3.5      & 1        & 3.5 \\
Glass2                     & 6        & 1.5      & 5        & 1.5      & 3        & 4   \\
Glass4                     & 6        & 1.5      & 5        & 3        & 1.5      & 4   \\
Haber                      & 5        & 6        & 1        & 3        & 4        & 2   \\
Monk1                      & 2        & 2        & 6        & 5        & 4        & 2   \\
Monk2                      & 5.5      & 4        & 5.5      & 3        & 2        & 1   \\
New-thyroid1               & 5        & 5        & 5        & 2        & 2        & 2   \\
Shuttle\_6-vs-2\_3         & 2.5      & 2.5      & 2.5      & 2.5      & 5.5      & 5.5 \\
Shuttle\_c0-vs-c4           & 3.5      & 3.5      & 3.5      & 3.5      & 3.5      & 3.5 \\
Transfusion                & 5        & 1        & 6        & 2        & 3        & 4   \\
Vehicle 1                  & 5        & 3        & 4        & 2        & 6        & 1   \\
Yeast\_0\_2\_5\_6-vs-3\_7\_8\_9 & 6        & 3        & 4        & 5        & 1        & 2   \\
Yeast\_0\_3\_5\_9-vs-7\_8     & 3        & 1.5      & 6        & 1.5      & 4.5      & 4.5 \\
Yeast1vs7                  & 3        & 4.5      & 1        & 4.5      & 6        & 2   \\
Yeast3                     & 3        & 6        & 4        & 5        & 2        & 1   \\
\hline
Average rank                  & 4.47 & 3.57 & 3.45 & 3.47 & \textbf{3.05} & \textbf{3}   \\
  \hline
\end{tabular}}}
\end{table}

\begin{table}[]
\centering
{\caption{Performance comparison on real world datasets with Gaussian kernel.}
\label{table:1}
 \resizebox{1.1\textwidth}{!}{
\begin{tabular}{lcccccc}
\hline
DATASETS &
  LSTSVM \cite{ARUNKUMAR20097535}&
  ELS-TSVM \cite{nasiri2014energy} &
  IFLSTSVM \cite{laxmi2022intuitionistic} &
  RELS-TSVM \cite{tanveer2016robust} &
  \begin{tabular}[c]{@{}l@{}}Proposed\\ IF-RELSTSVM\end{tabular} &
  \begin{tabular}[c]{@{}l@{}}Proposed\\ F-RELSTSVM\end{tabular} \\
 &
  AUC(\%), Time(s) &
  AUC(\%), Time(s) &
  AUC(\%), Time(s) &
  AUC(\%), Time(s) &
  AUC(\%), Time(s) &
  AUC(\%), Time(s) \\
  \hline
Abalone9-18 &
  80.49$,$ 0.04575 &
  82.19$,$ 0.11635 &
  79.56$,$ 0.11343 &
  \textbf{82.43}$,$ 0.05702 &
  73.53$,$ 0.08418 &
  82.19$,$ 0.14283 \\
Brwisconsin &
  95.94$,$ 0.0281 &
  97.38$,$ 0.01777 &
  97.43$,$ 0.05074 &
  \textbf{97.79}$,$ 0.05163 &
  97.43$,$ 0.04809 &
  97.43$,$ 0.05744 \\
Bupa or liver-disorders &
  57.39$,$ 0.00433 &
  55.61$,$ 0.00337 &
  60.46$,$ 0.01235 &
  \textbf{68.99}$,$ 0.00598 &
  61.52$,$ 0.01678 &
  62.58$,$ 0.01725 \\
Crossplane150 &
  \textbf{100}$,$ 0.00467 &
  \textbf{100}$,$ 0.00133 &
  98.21$,$ 0.0044 &
  \textbf{100}$,$ 0.00131 &
  \textbf{100}$,$ 0.00647 &
  \textbf{100}$,$ 0.01177 \\
Ecoli\_0\_1\_4\_6-vs-5 &
  \textbf{99.38}$,$ 0.00204 &
  \textbf{99.38}$,$ 0.00232 &
  98.15$,$ 0.00602 &
  98.15$,$ 0.00417 &
  98.77$,$ 0.00599 &
  98.77$,$ 0.00707 \\
Ecoli\_0\_1\_4\_7-vs-5\_6 &
  81.67$,$ 0.0057 &
  71.67$,$ 0.00293 &
  81.11$,$ 0.00802 &
  71.67$,$ 0.00596 &
  76.11$,$ 0.00825 &
  \textbf{86.11}$,$ 0.00883 \\
Ecoli\_0\_1-vs-2\_3\_5 &
  81.82$,$ 0.00322 &
  77.27$,$ 0.00179 &
  86.36$,$ 0.00582 &
  \textbf{90.10}$,$ 0.00315 &
  81.82$,$ 0.00498 &
  85.56$,$ 0.00517 \\
Ecoli\_0\_1-vs-5 &
  81.25$,$ 0.00176 &
  81.25$,$ 0.00192 &
  87.5$,$ 0.00455 &
  83.59$,$ 0.00291 &
  87.5$,$ 0.00578 &
  \textbf{91.41}$,$ 0.00546 \\
Ecoli\_0\_2\_6\_7-vs-3\_5 &
  70.6$,$ 0.00132 &
  71.43$,$ 0.00171 &
  \textbf{80.71}$,$ 0.00436 &
  71.43$,$ 0.00272 &
  79.05$,$ 0.00526 &
  \textbf{80.71}$,$ 0.00455 \\
Ecoli\_0\_3\_4\_7-vs-5\_6 &
  87.12$,$ 0.00199 &
  80.3$,$ 0.00203 &
  83.33$,$ 0.00407 &
  86.36$,$ 0.00312 &
  88.64$,$ 0.00591 &
  \textbf{91.67}$,$ 0.00482 \\
Ecoli01vs5 &
  98.41$,$ 0.00189 &
  99.21$,$ 0.00167 &
  97.62$,$ 0.00565 &
  \textbf{100}$,$ 0.00296 &
  98.41$,$ 0.00714 &
  98.41$,$ 0.00501 \\
Ecoli4 &
  96.24$,$ 0.00267 &
  85.71$,$ 0.00296 &
  92.86$,$ 0.00845 &
  95.7$,$ 0.00612 &
  \textbf{97.85}$,$ 0.01172 &
  96.77$,$ 0.00936 \\
Glass\_0\_1\_4\_6-vs-2 &
  44.96$,$ 0.00156 &
  44.96$,$ 0.00168 &
  57.46$,$ 0.00477 &
  62.72$,$ 0.00254 &
  37.72$,$ 0.00986 &
  \textbf{67.11}$,$ 0.00367 \\
Glass\_0\_1\_6-vs-5 &
  73.11$,$ 0.00121 &
  \textbf{99.06}$,$ 0.00142 &
  73.11$,$ 0.00479 &
  \textbf{99.06}$,$ 0.00216 &
  97.17$,$ 0.00484 &
  97.17$,$ 0.00326 \\
Glass\_0\_6-vs-5 &
  \textbf{100}$,$ 0.00122 &
  \textbf{100}$,$ 0.00059 &
  \textbf{100}$,$ 0.003 &
  \textbf{100}$,$ 0.00094 &
  \textbf{100}$,$ 0.00293 &
  \textbf{100}$,$ 0.00185 \\
Glass2 &
  61.49$,$ 0.00155 &
  48.56$,$ 0.00155 &
  58.33$,$ 0.00521 &
  51.44$,$ 0.00194 &
  \textbf{65.52}$,$ 0.00586 &
  54.02$,$ 0.00422 \\
Glass4 &
  \textbf{95.54}$,$ 0.00141 &
  77.68$,$ 0.00164 &
  81.25$,$ 0.00628 &
  84.82$,$ 0.00198 &
  82.14$,$ 0.00584 &
  91.07$,$ 0.00381 \\
Haber &
  67.17$,$ 0.00274 &
  \textbf{69.47}$,$ 0.00222 &
  62.28$,$ 0.00779 &
  69.37$,$ 0.0042 &
  68$,$ 0.0083 &
  69.41$,$ 0.00667 \\
Monk1 &
  46.62$,$ 0.0086 &
  50$,$ 0.01137 &
  40.37$,$ 0.02167 &
  50.25$,$ 0.01595 &
  47.66$,$ 0.02907 &
  \textbf{57.25}$,$ 0.02955 \\
Monk2 &
  \textbf{79.37}$,$ 0.02808 &
  77.38$,$ 0.02313 &
  71.69$,$ 0.02916 &
  77.38$,$ 0.03997 &
  71.69$,$ 0.04417 &
  77.78$,$ 0.07508 \\
New-thyroid1 &
  \textbf{93.75}$,$ 0.00154 &
  85.71$,$ 0.00152 &
  87.5$,$ 0.00521 &
  92.86$,$ 0.00265 &
  81.25$,$ 0.00495 &
  \textbf{93.75}$,$ 0.00414 \\
Shuttle\_6-vs-2\_3 &
  50$,$ 0.00143 &
  50$,$ 0.00138 &
  \textbf{98.51}$,$ 0.00433 &
  \textbf{98.51}$,$ 0.0029 &
  97.76$,$ 0.00497 &
  \textbf{98.51}$,$ 0.00443 \\
Shuttle\_c0-vs-c4 &
  \textbf{99.9}$,$ 0.20172 &
  99.8$,$ 0.16058 &
  \textbf{99.9}$,$ 0.31575 &
  99.8$,$ 0.4391 &
  \textbf{99.9}$,$ 0.4381 &
  \textbf{99.9}$,$ 0.52959 \\
Transfusion &
  56.85$,$ 0.02267 &
  55.53$,$ 0.05365 &
  57.9$,$ 0.07529 &
  54.33$,$ 0.23136 &
  \textbf{60.21}$,$ 0.04921 &
  55.19$,$ 0.04601 \\
Vehicle 1 &
  66.47$,$ 0.02946 &
  64.61$,$ 0.02369 &
  67.24$,$ 0.04773 &
  76.28$,$ 0.04773 &
  67.51$,$ 0.06212 &
  \textbf{76.55}$,$ 0.07874 \\
Yeast\_0\_2\_5\_6-vs-3\_7\_8\_9 &
  \textbf{76.67}$,$ 0.04745 &
  \textbf{76.67}$,$ 0.03281 &
  72.79$,$ 0.07591 &
  71.86$,$ 0.0727 &
  71.86$,$ 0.12351 &
  73.71$,$ 0.10853 \\
Yeast\_0\_3\_5\_9-vs-7\_8 &
  61.71$,$ 0.00651 &
  63.56$,$ 0.0048 &
  63.19$,$ 0.01574 &
  60.81$,$ 0.01086 &
  \textbf{65.42}$,$ 0.01965 &
  63.94$,$ 0.01868 \\
Yeast1vs7 &
  61.83$,$ 0.00536 &
  \textbf{73.09}$,$ 0.00477 &
  57.29$,$ 0.01699 &
  65.98$,$ 0.00859 &
  63.6$,$ 0.01888 &
  67.17$,$ 0.01519 \\
Yeast3 &
  94.47$,$ 0.12713 &
  91.95$,$ 0.07478 &
  93.65$,$ 0.16045 &
  91.67$,$ 0.23954 &
  \textbf{96.07}$,$ 0.24887 &
  95.95$,$ 0.32764 \\
  \hline
Average AUC(\%), Average time(s)&
  77.94$,$ 0.02045 &
  76.88$,$ 0.01923 &
  78.82$,$ 0.03545 &
  \textbf{81.15$,$ 0.04387} &
  79.8$,$ 0.04454 &
  \textbf{83.11}$,$ 0.05313 \\
\hline
\end{tabular}}}
\end{table}

\begin{table}[]
\centering{
\caption{Comparison on ranks based on AUC for real world datasets with Gaussian kernel.}
 \label{table:4}
\resizebox{1.1\textwidth}{!}{
\begin{tabular}{lcccccc}
\hline

DATASETS &
  LSTSVM \cite{ARUNKUMAR20097535}&
  ELS-TSVM \cite{nasiri2014energy} &
  IFLSTSVM \cite{laxmi2022intuitionistic} &
  RELS-TSVM \cite{tanveer2016robust} &
  \begin{tabular}[c]{@{}l@{}}Proposed\\ IF-RELSTSVM\end{tabular} &
  \begin{tabular}[c]{@{}l@{}}Proposed\\ F-RELSTSVM\end{tabular} \\
 \hline 
Abalone9-18                & 4        & 2.5      & 5        & 1        & 6        & 2.5      \\
Brwisconsin                & 6        & 5        & 3        & 1        & 3        & 3        \\
Bupa or liver-disorders    & 5        & 6        & 4        & 1        & 3        & 2        \\
Crossplane150              & 3        & 3        & 6        & 3        & 3        & 3        \\
Ecoli\_0\_1\_4\_6-vs-5       & 1.5      & 1.5      & 5.5      & 5.5      & 3.5      & 3.5      \\
Ecoli\_0\_1\_4\_7-vs-5\_6     & 2        & 5.5      & 3        & 5.5      & 4        & 1        \\
Ecoli\_0\_1-vs-2\_3\_5       & 4.5      & 6        & 2        & 1        & 4.5      & 3        \\
Ecoli\_0\_1-vs-5           & 5.5      & 5.5      & 2.5      & 4        & 2.5      & 1        \\
Ecoli\_0\_2\_6\_7-vs-3\_5     & 6        & 4.5      & 1.5      & 4.5      & 3        & 1.5      \\
Ecoli\_0\_3\_4\_7-vs-5\_6     & 3        & 6        & 5        & 4        & 2        & 1        \\
Ecoli01vs5                 & 4        & 2        & 6        & 1        & 4        & 4        \\
Ecoli4                     & 3        & 6        & 5        & 4        & 1        & 2        \\
Glass\_0\_1\_4\_6-vs-2       & 4.5      & 4.5      & 3        & 2        & 6        & 1        \\
Glass\_0\_1\_6-vs-5         & 5.5      & 1.5      & 5.5      & 1.5      & 3.5      & 3.5      \\
Glass\_0\_6-vs-5           & 3.5      & 3.5      & 3.5      & 3.5      & 3.5      & 3.5      \\
Glass2                     & 2        & 6        & 3        & 5        & 1        & 4        \\
Glass4                     & 1        & 6        & 5        & 3        & 4        & 2        \\
Haber                      & 5        & 1        & 6        & 3        & 4        & 2        \\
Monk1                      & 5        & 3        & 6        & 2        & 4        & 1        \\
Monk2                      & 1        & 3.5      & 5.5      & 3.5      & 5.5      & 2        \\
New-thyroid1               & 1.5      & 5        & 4        & 3        & 6        & 1.5      \\
Shuttle\_6-vs-2\_3         & 5.5      & 5.5      & 2        & 2        & 4        & 2        \\
Shuttle\_c0-vs-c4           & 2.5      & 5.5      & 2.5      & 5.5      & 2.5      & 2.5      \\
Transfusion                & 3        & 4        & 2        & 6        & 1        & 5        \\
Vehicle 1                  & 5        & 6        & 4        & 2        & 3        & 1        \\
Yeast\_0\_2\_5\_6-vs-3\_7\_8\_9 & 1.5      & 1.5      & 4        & 6        & 5        & 3        \\
Yeast\_0\_3\_5\_9-vs-7\_8     & 5        & 3        & 4        & 6        & 1        & 2        \\
Yeast1vs7                  & 5        & 1        & 6        & 3        & 4        & 2        \\
Yeast3                     & 3        & 5        & 4        & 6        & 1        & 2        \\
\hline
Average rank                    & 3.69 & 4.10 & 4.09 & \textbf{3.39} & \textbf{3.39} & \textbf{2.33} \\
  \hline

\hline
\end{tabular}}}
\end{table}

\subsection{Noise and synthetic datasets}
To assess the effectiveness of the proposed F-RELSTSVM and IF-RELSTSVM methods, we conducted experiments using synthetic datasets. Specifically, we utilized noisy synthetic datasets in our analysis. These datasets are obtained from the KEEL imbalanced datasets repository 
\cite{derrac2015keel, napierala2010learning}. The datasets consist of two classes with data points randomly and uniformly distributed in a two-dimensional space (both attributes being real-valued). The noisy datasets used in our experiments were identified with different disturbance ratios of $30\%, 50\%,$ and $60\% $
\cite{napierala2010learning}. From Table \ref{table:6}, it is evident that the proposed approach achieves the lowest rank with the highest average AUC with linear kernel in F-RELSTSVM.
We also conducted experiments on the crossplane (XOR) dataset \cite{shao2011improvements}, which involved generating datasets with varying sample sizes and imbalance ratios. The datasets were created using randomized values in the equation of a line, $y = mx + b$. For the negative class, we selected the slope parameters $m=0.7$ and intercept $b=0.1$. Conversely, for the positive class, we chose parameter values as $m=-0.6$ and $b=1$.

\subsubsection{Statistical analysis on synthetic datasets}
To assess the statistical significance of the proposed F-RELSTSVM and IF-RELSTSVM, we employ the Friedman test along with the corresponding post-hoc test. This analysis is performed on $22$ noisy and synthetic binary class datasets with Gaussian kernel. Under the null hypothesis, we assume that all algorithms perform equivalently. The Friedman statistic is computed based on the ranks of the AUC values presented in Table (\ref{table:7}).
\begin{align}
\chi_F^2=&\frac{12 \times 22}{6(6+1)}\left[3.36^2+4.48^2+5.09^2+2.84^2+2.7^2+2.52^2-\frac{6(7)^2}{4}\right]\\=&34.38, \nonumber \\
F_F=&\frac{(N-1) \chi_F^2}{N(k-1)-\chi_F^2}=\frac{(22-1) \times 34.38 }{22(6-1)-34.38}=9.54 .
\end{align}
\begin{figure}
  \includegraphics[width=5.0in,height=2.5in]{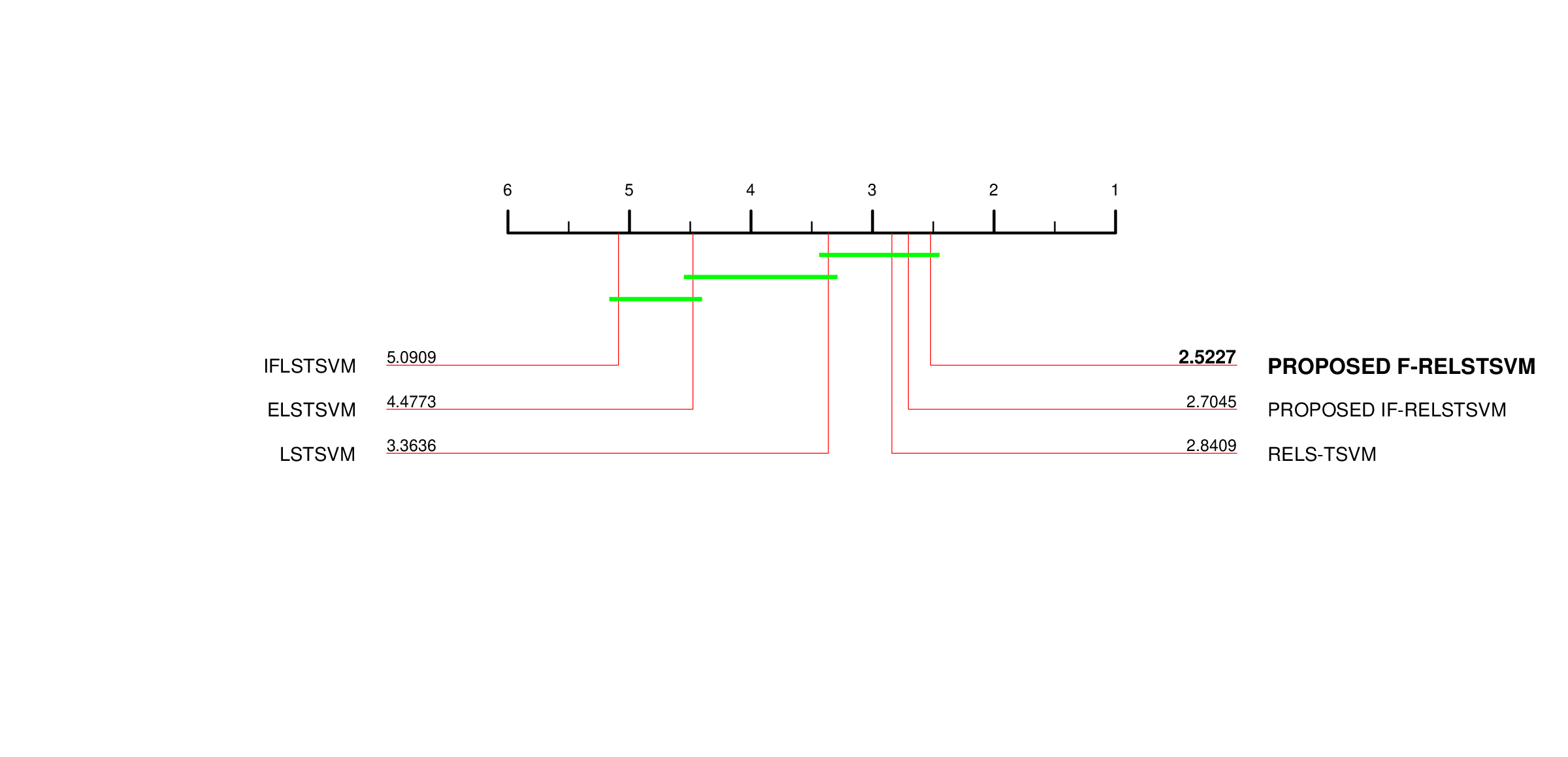}
    \caption{Critical difference diagram with Gaussian kernel on noise and synthetic datasets}
     \label{fig:CD3}
\end{figure}
The value $F_F$ is distributed according to the $F$-distribution with degrees of freedom $(5, 105)$, as there are six algorithms and $22$ datasets involved. Considering a significance level of $\alpha$ = $0.05$, the critical value for $F(5, 105)$ is $2.30$. Since the computed value of $F_F$ is $10.99$, which exceeds the critical value of $2.30$, we reject the null hypothesis. Additionally, we conducted the Nemenyi post-hoc test to perform pairwise comparisons between the algorithms. The $CD$ is calculated at a significance level of $p = 0.10$ and should exhibit a minimum difference of $1.46$ to indicate a significant distinction between the algorithms, clearly seen in Fig. (\ref{fig:CD3}).

\begin{table}[]
\centering

\caption{Performance comparison on synthetic datasets with linear kernel.}

 \label{table:6}
\resizebox{1.1\textwidth}{!}{
\begin{tabular}{lcccccc}
\hline

DATASETS &
  LSTSVM \cite{ARUNKUMAR20097535}&
  ELS-TSVM \cite{nasiri2014energy} &
  IFLSTSVM \cite{laxmi2022intuitionistic} &
  RELS-TSVM \cite{tanveer2016robust} &
  \begin{tabular}[c]{@{}l@{}}Proposed\\ IF-RELSTSVM\end{tabular} &
  \begin{tabular}[c]{@{}l@{}}Proposed\\ F-RELSTSVM\end{tabular} \\
&
  AUC(\%), Time(s) &
  AUC(\%), Time(s) &
  AUC(\%), Time(s) &
  AUC(\%), Time(s) &
  AUC(\%), Time(s) &
  AUC(\%), Time(s) \\
  \hline
03subcl5-600-5-0-BI &
  50$,$ 0.00386 &
  50$,$ 0.00123 &
  59.12$,$ 0.01982 &
  64.19$,$ 0.00378 &
  59.71$,$ 0.02455 &
  \textbf{64.44}$,$ 0.01371 \\
03subcl5-600-5-30-BI &
  58.57$,$ 0.00266 &
  59.92$,$ 0.00032 &
  51.35$,$ 0.01596 &
  \textbf{63.01}$,$ 0.00277 &
  61.66$,$ 0.01555 &
  58.61$,$ 0.01092 \\
03subcl5-600-5-50-BI &
  57.77$,$ 0.0028 &
  53.38$,$ 0.00036 &
  50$,$ 0.0078 &
  58.53$,$ 0.00269 &
  \textbf{61.87}$,$ 0.0142 &
  60.22$,$ 0.00941 \\
03subcl5-600-5-70-BI &
  50$,$ 0.00507 &
  50$,$ 0.00159 &
  49.03$,$ 0.02128 &
  55.74$,$ 0.00642 &
  56$,$ 0.02325 &
  \textbf{63.85}$,$ 0.01488 \\
03subcl5-800-7-0-BI &
  55.99$,$ 0.00657 &
  59.41$,$ 0.00015 &
  55.3$,$ 0.01383 &
  68.66$,$ 0.00913 &
  \textbf{69.02}$,$ 0.01652 &
  59.69$,$ 0.01489 \\
03subcl5-800-7-30-BI &
  47.32$,$ 0.00671 &
  45.11$,$ 0.00017 &
  51.08$,$ 0.0145 &
  \textbf{60.3}$,$ 0.00497 &
  58.69$,$ 0.01904 &
  58.36$,$ 0.01587 \\
03subcl5-800-7-50-BI &
  55.2$,$ 0.00497 &
  55.49$,$ 0.0002 &
  47.92$,$ 0.01476 &
  \textbf{59.15$,$ 0.00512} &
  58.69$,$ 0.01724 &
  57.66$,$ 0.01893 \\
03subcl5-800-7-60-BI &
  53.23$,$ 0.00528 &
  52.3$,$ 0.00019 &
  53.69$,$ 0.01017 &
  59.91$,$ 0.005 &
  \textbf{66.95}$,$ 0.02366 &
  56.02$,$ 0.01495 \\
03subcl5-800-7-70-BI &
  48.08$,$ 0.00501 &
  50.25$,$ 0.00019 &
  50$,$ 0.01387 &
  \textbf{54.11}$,$ 0.00527 &
  52.99$,$ 0.01699 &
  53.09$,$ 0.01518 \\
04clover5z-600-5-30-BI &
  63.3$,$ 0.00246 &
  59.12$,$ 0.00018 &
  65.12$,$ 0.00625 &
  58.23$,$ 0.00191 &
  \textbf{67.69}$,$ 0.00864 &
  61.95$,$ 0.00473 \\
04clover5z-600-5-50-BI &
  51.69$,$ 0.00298 &
  50$,$ 0.00015 &
  58.45$,$ 0.00636 &
  54.05$,$ 0.00233 &
  \textbf{60.43}$,$ 0.0091 &
  53$,$ 0.00502 \\
04clover5z-600-5-60-BI &
  53.38$,$ 0.00218 &
  50$,$ 0.00017 &
  51.01$,$ 0.00657 &
  50$,$ 0.00208 &
  47$,$ 0.01381 &
  \textbf{64.57}$,$ 0.0049 \\
04clover5z-800-7-30-BI &
 50$,$ 0.00501 &
  51.96$,$ 0.00017 &
  58.98$,$ 0.01124 &
  51.96$,$ 0.00494 &
  \textbf{67.76}$,$ 0.03004 &
  54.8$,$ 0.01835 \\
04clover5z-800-7-50-BI &
  57.76$,$ 0.00487 &
  55.66$,$ 0.00017 &
  \textbf{62.33}$,$ 0.0104 &
  57.5$,$ 0.00508 &
  61.48$,$ 0.02089 &
  56.61$,$ 0.01226 \\
04clover5z-800-7-60-BI &
  59.91$,$ 0.0054 &
  55.76$,$ 0.00013 &
  \textbf{63.71}$,$ 0.01173 &
  50$,$ 0.00513 &
  45.17$,$ 0.02329 &
  40.13$,$ 0.0148 \\
04clover5z-800-7-70-BI &
  50$,$ 0.00526 &
  50$,$ 0.00018 &
  \textbf{67.83}$,$ 0.01101 &
  50$,$ 0.00553 &
  58.59$,$ 0.02075 &
  55.85$,$ 0.02297 \\
Paw02a-600-5-0-BI &
  50$,$ 0.00217 &
  50$,$ 0.00015 &
  50$,$ 0.00813 &
  58.78$,$ 0.00211 &
  \textbf{61.87}$,$ 0.01249 &
  44.26$,$ 0.00886 \\
Paw02a-600-5-30-BI &
  58.45$,$ 0.00226 &
  57.77$,$ 0.00016 &
  \textbf{63.34$,$ 0.00659} &
  56.33$,$ 0.00216 &
  54.48$,$ 0.01195 &
  62.25$,$ 0.00962 \\
Paw02a-800-7-0-BI &
  51.15$,$ 0.00483 &
  61.42$,$ 0.00015 &
  50$,$ 0.01007 &
  61.88$,$ 0.00518 &
  57.66$,$ 0.02142 &
  \textbf{68.03}$,$ 0.01132 \\
Paw02a-800-7-30-BI &
  57.37$,$ 0.00467 &
  58.36$,$ 0.00015 &
  \textbf{61.98}$,$ 0.00988 &
  61.78$,$ 0.00469 &
  50$,$ 0.02148 &
  61.34$,$ 0.012 \\
Crossplane\_450 &
  99.58$,$ 0.00205 &
  99.58$,$ 0.00027 &
  \textbf{100}$,$ 0.01072 &
  99.58$,$ 0.00123 &
  96.45$,$ 0.01278 &
  \textbf{100}$,$ 0.00851 \\
Crossplane\_500 &
  \textbf{100}$,$ 0.00216 &
  \textbf{100}$,$ 0.00031 &
  \textbf{100}$,$ 0.00758 &
  \textbf{100}$,$ 0.0019 &
  \textbf{100}$,$ 0.01092 &
  \textbf{100}$,$ 0.00734 \\
  \hline
Average AUC(\%), Average time(s) &
  58.12$,$ 0.00405 &
  57.98$,$ 0.00031 &
  60.01$,$ 0.0113 &
  61.53$,$ 0.00406 &
  \textbf{62.46$,$ 0.01766} &
  61.58$,$ 0.01225\\
  \hline
\end{tabular}}
\end{table}

\begin{table}[]
\centering
{
\caption{Rank comparison on synthetic datasets with linear kernel.}

 \label{table:8}
\resizebox{1.1\textwidth}{!}{
\begin{tabular}{lcccccc}
\hline
DATASETS &
  LSTSVM \cite{ARUNKUMAR20097535}&
  ELS-TSVM \cite{nasiri2014energy} &
  IFLSTSVM \cite{laxmi2022intuitionistic} &
  RELS-TSVM \cite{tanveer2016robust} &
  \begin{tabular}[c]{@{}l@{}}Proposed\\ IF-RELSTSVM\end{tabular} &
  \begin{tabular}[c]{@{}l@{}}Proposed\\ F-RELSTSVM\end{tabular} \\
 \hline 
03subcl5-600-5-0-BI    & 5.5  & 5.5  & 4    & 2    & 3   & 1    \\
03subcl5-600-5-30-BI   & 5    & 3    & 6    & 1    & 2   & 4    \\
03subcl5-600-5-50-BI   & 4    & 5    & 6    & 3    & 1   & 2    \\
03subcl5-600-5-70-BI   & 4.5  & 4.5  & 6    & 3    & 2   & 1    \\
03subcl5-800-7-0-BI    & 5    & 4    & 6    & 2    & 1   & 3    \\
03subcl5-800-7-30-BI   & 5    & 6    & 4    & 1    & 2   & 3    \\
03subcl5-800-7-50-BI   & 5    & 4    & 6    & 1    & 2   & 3    \\
03subcl5-800-7-60-BI   & 5    & 6    & 4    & 2    & 1   & 3    \\
03subcl5-800-7-70-BI   & 6    & 4    & 5    & 1    & 3   & 2    \\
04clover5z-600-5-30-BI & 3    & 5    & 2    & 6    & 1   & 4    \\
04clover5z-600-5-50-BI & 5    & 6    & 2    & 3    & 1   & 4    \\
04clover5z-600-5-60-BI & 2    & 4.5  & 3    & 4.5  & 6   & 1    \\
04clover5z-800-7-30-BI & 6    & 4.5  & 2    & 4.5  & 1   & 3    \\
04clover5z-800-7-50-BI & 3    & 6    & 1    & 4    & 2   & 5    \\
04clover5z-800-7-60-BI & 2    & 3    & 1    & 4    & 5   & 6    \\
04clover5z-800-7-70-BI & 5    & 5    & 1    & 5    & 2   & 3    \\
Paw02a-600-5-0-BI      & 4    & 4    & 4    & 2    & 1   & 6    \\
Paw02a-600-5-30-BI     & 3    & 4    & 1    & 5    & 6   & 2    \\
Paw02a-800-7-0-BI      & 5    & 3    & 6    & 2    & 4   & 1    \\
Paw02a-800-7-30-BI     & 5    & 4    & 1    & 2    & 6   & 3    \\
Crossplane\_450        & 4    & 4    & 1.5  & 4    & 6   & 1.5  \\
Crossplane\_500        & 3.5  & 3.5  & 3.5  & 3.5  & 3.5 & 3.5  \\
\hline
Average rank        & 4.34 & 4.48 & 3.45 & 2.98 & 2.8 & 2.95 \\
  \hline

\hline
\end{tabular}}}
\end{table}

\begin{table}[]
\centering

\caption{Performance comparison on synthetic datasets with Gaussian kernel.}

 \label{table:5}
\resizebox{1.1\textwidth}{!}{
\begin{tabular}{lcccccc}
\hline

DATASETS &
  LSTSVM \cite{ARUNKUMAR20097535}&
  ELS-TSVM \cite{nasiri2014energy} &
  IFLSTSVM \cite{laxmi2022intuitionistic} &
  RELS-TSVM \cite{tanveer2016robust} &
  \begin{tabular}[c]{@{}l@{}}Proposed\\ IF-RELSTSVM\end{tabular} &
  \begin{tabular}[c]{@{}l@{}}Proposed\\ F-RELSTSVM\end{tabular} \\  
 &
  AUC(\%), Time(s) &
  AUC(\%), Time(s) &
  AUC(\%), Time(s) &
  AUC(\%), Time(s) &
  AUC(\%), Time(s) &
  AUC(\%), Time(s) \\
  \hline
03subcl5-600-5-0-BI &
  83.36$,$ 0.03875 &
  83.36$,$ 0.03056 &
  73.99$,$ 0.05803 &
  92.95$,$ 0.06509 &
  93.03$,$ 0.09131 &
  \textbf{93.62}$,$ 0.10386 \\
03subcl5-600-5-30-BI &
  75$,$ 0.04434 &
  77.58$,$ 0.02795 &
  66.39$,$ 0.04901 &
  74.96$,$ 0.06601 &
  \textbf{81.97}$,$ 0.06768 &
  75$,$ 0.09756 \\
03subcl5-600-5-50-BI &
  71.83$,$ 0.03484 &
  67.69$,$ 0.02306 &
  58.57$,$ 0.04374 &
  73.02$,$ 0.056 &
  \textbf{75.93}$,$ 0.07242 &
  75.8$,$ 0.09662 \\
03subcl5-600-5-70-BI &
  61.7$,$ 0.037 &
  68.24$,$ 0.02766 &
  61.36$,$ 0.04807 &
  73.99$,$ 0.05576 &
  69.81$,$ 0.07851 &
  \textbf{74.7}$,$ 0.10413 \\
03subcl5-800-7-0-BI &
  83.86$,$ 0.0679 &
  79.97$,$ 0.05103 &
  71.28$,$ 0.07249 &
  92.43$,$ 0.11963 &
  \textbf{93.12}$,$ 0.14713 &
  92.2$,$ 0.18821 \\
03subcl5-800-7-30-BI &
  \textbf{81.4}$,$ 0.07534 &
  64.96$,$ 0.05484 &
  64.63$,$ 0.08307 &
  74.87$,$ 0.13442 &
  73.85$,$ 0.15249 &
  74.18$,$ 0.19493 \\
03subcl5-800-7-50-BI &
  70.7$,$ 0.06712 &
  66.64$,$ 0.0558 &
  53.76$,$ 0.07817 &
  \textbf{76.92}$,$ 0.11743 &
  74.31$,$ 0.14087 &
  75.54$,$ 0.18732 \\
03subcl5-800-7-60-BI &
  74.51$,$ 0.06824 &
  57.74$,$ 0.05845 &
  65.65$,$ 0.07365 &
  76.63$,$ 0.12002 &
  71.62$,$ 0.13055 &
  \textbf{78.24}$,$ 0.19097 \\
03subcl5-800-7-70-BI &
  63.41$,$ 0.06906 &
  59.46$,$ 0.05118 &
  51.35$,$ 0.06581 &
  69.58$,$ 0.12281 &
  62.85$,$ 0.12686 &
  \textbf{73.82}$,$ 0.19475 \\
04clover5z-600-5-30-BI &
  \textbf{92.02}$,$ 0.02988 &
  78.63$,$ 0.02694 &
  76.31$,$ 0.03289 &
  85.14$,$ 0.0527 &
  84.71$,$ 0.05968 &
  89.53$,$ 0.08698 \\
04clover5z-600-5-50-BI &
  67.4$,$ 0.03458 &
  73.48$,$ 0.02866 &
  64.82$,$ 0.03425 &
  78.63$,$ 0.0557 &
  76.22$,$ 0.06151 &
  \textbf{79.43}$,$ 0.08919 \\
04clover5z-600-5-60-BI &
  \textbf{79.05}$,$ 0.03482 &
  54.39$,$ 0.02666 &
  60.47$,$ 0.03473 &
  68.67$,$ 0.05609 &
  62.8$,$ 0.06062 &
  77.91$,$ 0.09168 \\
04clover5z-800-7-30-BI &
  \textbf{78.43}$,$ 0.07276 &
  67.82$,$ 0.05407 &
  68.87$,$ 0.06971 &
  75.34$,$ 0.12518 &
  78.23$,$ 0.1621 &
  69.3$,$ 0.19545 \\
04clover5z-800-7-50-BI &
  \textbf{73.56}$,$ 0.07542 &
  57.74$,$ 0.05455 &
  62.68$,$ 0.07012 &
  66.18$,$ 0.13582 &
  71.12$,$ 0.14211 &
  69.84$,$ 0.20474 \\
04clover5z-800-7-60-BI &
  62.19$,$ 0.06764 &
  56.03$,$ 0.05498 &
  58.34$,$ 0.07491 &
  65.49$,$ 0.12321 &
  67$,$ 0.13505 &
  \textbf{69.61}$,$ 0.19664 \\
04clover5z-800-7-70-BI &
  63.21$,$ 0.06764 &
  50.76$,$ 0.05412 &
  56.39$,$ 0.07267 &
  59.76$,$ 0.12286 &
  \textbf{65.95}$,$ 0.13424 &
  64.34$,$ 0.19374 \\
Paw02a-600-5-0-BI &
  96.41$,$ 0.02145 &
  \textbf{98.65}$,$ 0.02048 &
  93.96$,$ 0.03468 &
  94.93$,$ 0.04034 &
  96.96$,$ 0.04603 &
  96.96$,$ 0.06236 \\
Paw02a-600-5-30-BI &
  70.35$,$ 0.02495 &
  83.78$,$ 0.02155 &
  \textbf{84.59}$,$ 0.03316 &
  82.73$,$ 0.04146 &
  84.33$,$ 0.05043 &
  81.97$,$ 0.06953 \\
Paw02a-800-7-0-BI &
  \textbf{96.08}$,$ 0.05678 &
  94.73$,$ 0.04067 &
  95.42$,$ 0.06393 &
  95.75$,$ 0.08512 &
  95.29$,$ 0.10425 &
  95.75$,$ 0.13335 \\
Paw02a-800-7-30-BI &
  79.06$,$ 0.05723 &
  79.38$,$ 0.04313 &
  72.3$,$ 0.0627 &
  84.16$,$ 0.08904 &
  \textbf{88.31}$,$ 0.1052 &
  76.29$,$ 0.14396 \\
Crossplane\_450 & 99.58$,$ 0.00588 & \textbf{100}$,$ 0.00558 & \textbf{100}$,$ 0.01774 & \textbf{100}$,$ 0.0126  & 99.58$,$ 0.0241         & 99.58$,$ 0.01883        \\
Crossplane\_500 & 96.88$,$ 0.01796 & \textbf{100}$,$ 0.00677 & \textbf{100}$,$ 0.01843 & \textbf{100}$,$ 0.01713 & \textbf{100}$,$ 0.02427 & \textbf{100}$,$ 0.02325 \\
\hline
Average AUC(\%), Average time(s) &
  78.18$,$ 0.04862 &
  73.68$,$ 0.03721 &
  70.96$,$ 0.05418 &
  80.1$,$ 0.08247 &
  80.32$,$ 0.09625 &
\textbf{  81.07}$,$ 0.13037\\
  \hline

\hline
\end{tabular}}
\end{table}

\begin{table}[]
\centering{
\caption{Rank comparison on synthetic datasets with Gaussian  kernel.}

 \label{table:7}
\resizebox{1.1\textwidth}{!}{
\begin{tabular}{lcccccc}
\hline
DATASETS &
  LSTSVM \cite{ARUNKUMAR20097535}&
  ELS-TSVM \cite{nasiri2014energy} &
  IFLSTSVM \cite{laxmi2022intuitionistic} &
  RELS-TSVM \cite{tanveer2016robust} &
  \begin{tabular}[c]{@{}l@{}}Proposed\\ IF-RELSTSVM\end{tabular} &
  \begin{tabular}[c]{@{}l@{}}Proposed\\ F-RELSTSVM\end{tabular} \\
 \hline 
03subcl5-600-5-0-BI    & 4.5  & 4.5  & 6    & 3    & 2   & 1    \\
03subcl5-600-5-30-BI   & 3.5  & 2    & 6    & 5    & 1   & 3.5  \\
03subcl5-600-5-50-BI   & 4    & 5    & 6    & 3    & 1   & 2    \\
03subcl5-600-5-70-BI   & 5    & 4    & 6    & 2    & 3   & 1    \\
03subcl5-800-7-0-BI    & 4    & 5    & 6    & 2    & 1   & 3    \\
03subcl5-800-7-30-BI   & 1    & 5    & 6    & 2    & 4   & 3    \\
03subcl5-800-7-50-BI   & 4    & 5    & 6    & 1    & 3   & 2    \\
03subcl5-800-7-60-BI   & 3    & 6    & 5    & 2    & 4   & 1    \\
03subcl5-800-7-70-BI   & 3    & 5    & 6    & 2    & 4   & 1    \\
04clover5z-600-5-30-BI & 1    & 5    & 6    & 3    & 4   & 2    \\
04clover5z-600-5-50-BI & 5    & 4    & 6    & 2    & 3   & 1    \\
04clover5z-600-5-60-BI & 1    & 6    & 5    & 3    & 4   & 2    \\
04clover5z-800-7-30-BI & 1    & 6    & 5    & 3    & 2   & 4    \\
04clover5z-800-7-50-BI & 1    & 6    & 5    & 4    & 2   & 3    \\
04clover5z-800-7-60-BI & 4    & 6    & 5    & 3    & 2   & 1    \\
04clover5z-800-7-70-BI & 3    & 6    & 5    & 4    & 1   & 2    \\
Paw02a-600-5-0-BI      & 4    & 1    & 6    & 5    & 2.5 & 2.5  \\
Paw02a-600-5-30-BI     & 6    & 3    & 1    & 4    & 2   & 5    \\
Paw02a-800-7-0-BI      & 1    & 6    & 4    & 2.5  & 5   & 2.5  \\
Paw02a-800-7-30-BI     & 4    & 3    & 6    & 2    & 1   & 5    \\
Crossplane\_450        & 5    & 2    & 2    & 2    & 5   & 5    \\
Crossplane\_500        & 6    & 3    & 3    & 3    & 3   & 3    \\
\hline
Average rank        & 3.36 & 4.48 & 5.09 & 2.84 & 2.7 & 2.52 \\
  \hline

\hline
\end{tabular}}}
\end{table}

\subsection{Influence of energy parameters $E_1$ and $E_2$ on proposed models}

\begin{figure}
\centering
\begin{subfigure}{.5\textwidth}
  \centering
  \includegraphics[width=1.0\linewidth]{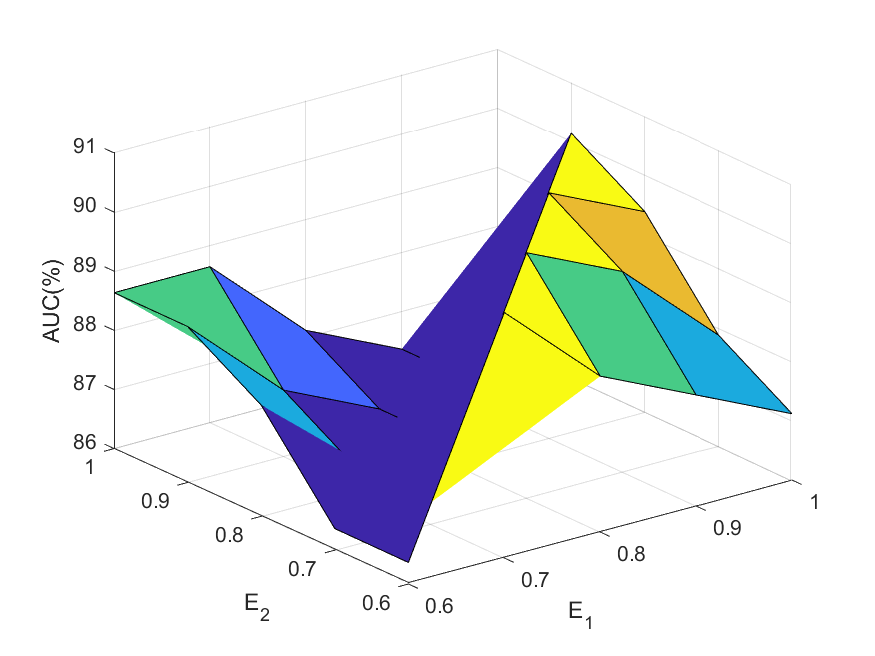}
  \caption{Ecoli-0-3-4-7$\_$vs$\_$5-6}
  \label{fig:sen1}
\end{subfigure}%
\begin{subfigure}{.5\textwidth}
  \centering
  \includegraphics[width=1.0\linewidth]{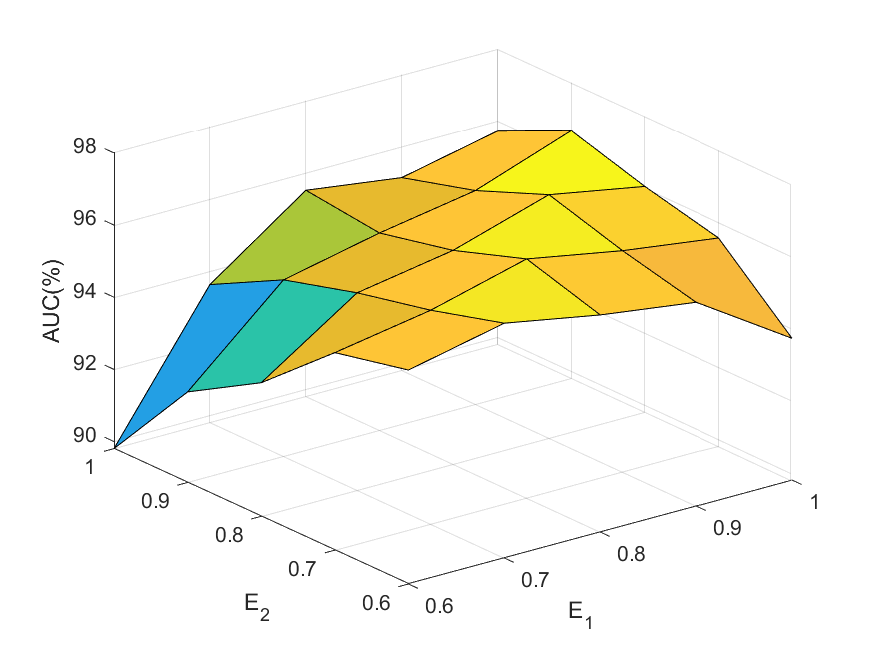}
  \caption{Yeast3}
  \label{fig:sen2}
\end{subfigure}
\caption{Performance of IF-RELSTSVM on energy parameters ($E_1$, $E_2$)
for Ecoli-0-3-4-7$\_$vs$\_$5-6 and Yeast3 datasets}
\begin{subfigure}{.5\textwidth}
  \centering
  \includegraphics[width=1.0\linewidth]{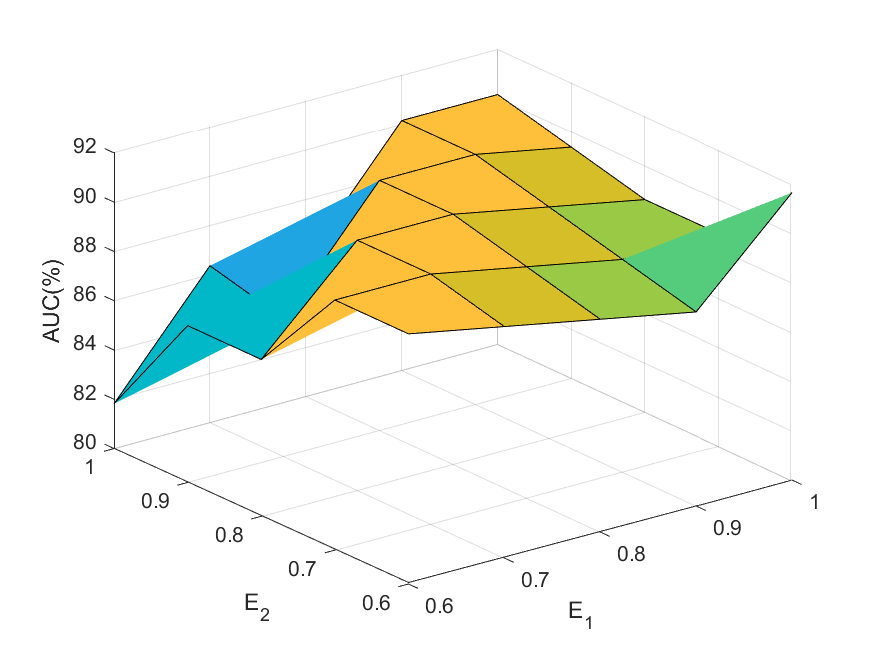}
  \caption{Ecoli-0-3-4-7$\_$vs$\_$5-6}
  \label{fig:sen3}
\end{subfigure}%
\begin{subfigure}{.5\textwidth}
  \centering
  \includegraphics[width=1.0\linewidth]{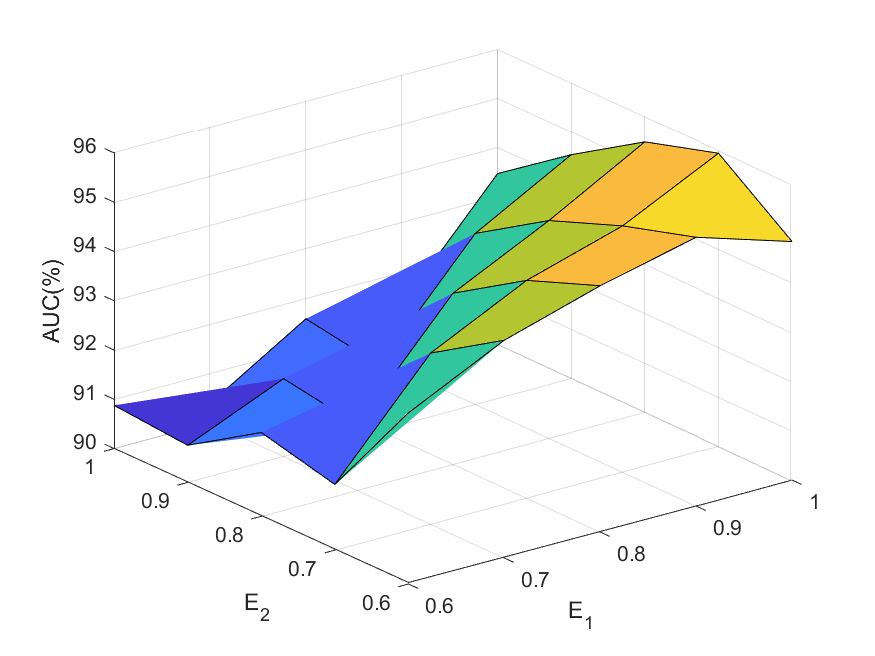}
  \caption{Yeast3}
  \label{fig:sen4}
\end{subfigure}
\caption{Performance of F-RELSTSVM on energy parameters ($E_1$, $E_2$)
for Ecoli-0-3-4-7$\_$vs$\_$5-6 and Yeast3 datasets}
\label{fig:test}
\end{figure}

To examine the influence of energy parameters $E_1$ and $E_2$ on the performance of our IF-RELSTSVM and F-RELSTSVM, we conducted experiments using the Ecoli-0-3-4-7$\_$vs$\_$5-6 and Yeast3 datasets. The hyperplane must be positioned at a distance of unity from points of other classes in LSTSVM, increasing its sensitivity to outliers. However, our IF-RELSTSVM and F-RELSTSVM introduce energy terms, $E_1$ and $E_2$ for each hyperplane similar to ELS-TSVM. By selecting suitable values for these parameters through grid search within the range [0.6, 0.7, 0.8, 0.9, 1.0], we can reduce the sensitivity of the classifier to noise, thereby improving its effectiveness and robustness. The energy parameters $E_1$ and $E_2$ are chosen such that the certainty of sample classification in either class is equal. If the $E_1/E_2$ ratio is large, it indicates a higher certainty for the sample to belong to one class over the other, and vice versa. Figs. (~\ref{fig:sen1}) and (\ref{fig:sen2}) showcase the impact of energy parameters $E_1$ and $E_2$ for classifying the Ecoli-0-3-4-7$\_$vs$\_$5-6 and Yeast3 datasets. Fig. (\ref{fig:sen1}) illustrates that IF-RELSTSVM performs better on the Ecoli-0-3-4-7$\_$vs$\_$5-6 dataset when $E_1$ and $E_2$ values are comparable. Similarly, Fig. (\ref{fig:sen2}) demonstrates that the proposed IF-RELSTSVM achieves better performance on Yeast3 with higher $E_1$ values and lower $E_2$ values, indicating the classifier's ability to adjust and reduce sample sensitivity. Likewise, Fig. (\ref{fig:sen3}) and (\ref{fig:sen4}) reveal improved performance on both the Ecoli-0-3-4-7$\_$vs$\_$5-6 and Yeast3 datasets respectively, with higher $E_1$ values and lower $E_2$ values. Notably, the performance of the F-RELSTSVM fluctuates as $E_1$ and $E_2$ values vary, reflecting the adaptive nature of the hyperplane, and its tendency to optimize energy values for better noise handling. 

\section{Credit card fraud detection}

To show the application of the proposed algorithms on real world problems, we take the credit card fraud detection dataset \cite{dal2017credit,lebichot2020deep} of large size and with a high imbalance ratio.

In the preprocessing, we performed PCA transformation on the features. Therefore, the datasets contain features from V1, V2,..., V28 as the principal components. The time and amount features are excluded from the PCA analysis. We include the amount feature later as the amount feature is used for dependent cost-sensitive learning. As datasets are large so we take datasets of different sizes viz. 10K, 20K, 30K, 40K, and 50K. The training size is 60\% of total samples, and chosen fixed value for $C_1, C_2, C_3, C_4$ as $1$, and $E_1$ and $E_2$ is set as 0.8 using linear kernel. The proposed IF-RELSTSVM and F-RELSTSVM outperform on credit card fraud detection dataset as we can see in Table \ref{table:9}. The average AUC of proposed IF-RELSTSVM and F-RELSTSVM are $89.82\%$ and $90.84\%$ respectively, followed by LSTSVM with $77.77\%$, IFLSTSVM with $77.37\%$, and both ELS-TSVM and RELS-TSVM achieve $77.21\%$. This demonstrates that our proposed algorithms IF-RELSTSVM and F-RELSTSVM perform better than the baseline algorithms.

\begin{table*}[]
\centering

\caption{Performance on credit card fraud detection.}

 \label{table:9}
\resizebox{1\textwidth}{!}{
\begin{tabular}{lcccccc}
\hline
DATASETS &
  LSTSVM \cite{ARUNKUMAR20097535}&
  ELS-TSVM \cite{nasiri2014energy} &
  IFLSTSVM \cite{laxmi2022intuitionistic} &
  RELS-TSVM \cite{tanveer2016robust} &
  \begin{tabular}[c]{@{}l@{}}Proposed\\ IF-RELSTSVM\end{tabular} &
  \begin{tabular}[c]{@{}l@{}}Proposed\\ F-RELSTSVM\end{tabular} \\
 \hline 
Creditcard\_10K & 85.42$,$ 0.50347  & 85.42$,$ 0.09123 & 85.42$,$ 1.02423  & 85.42$,$ 0.68781 & 89.35$,$ 1.85272  & \textbf{91.56}$,$ 14.6528  \\
Creditcard\_20K & 74.48$,$ 2.2466   & 74.48$,$ 0.00341 & 74.48$,$ 7.81414  & 74.48$,$ 3.58598 & 91.69$,$ 12.6311  & \textbf{92.79}$,$ 9.42395  \\
Creditcard\_30K & 74.95$,$ 6.27077  & 72.97$,$ 0.0044  & 72.97$,$ 16.6391  & 72.97$,$ 10.5538 & \textbf{86.94}$,$ 32.5731  & 85.43$,$ 27.1823  \\
Creditcard\_40K & 75.38$,$ 16.9848  & 74.56$,$ 0.00462 & 75.38$,$ 28.8133  & 74.56$,$ 29.7912 & 88.69$,$ 93.1857  & \textbf{90.51}$,$ 80.3976  \\
Creditcard\_50K & 78.61$,$ 48.0747  & 78.62$,$ 0.0113  & 78.61$,$ 86.9384  & 78.62$,$ 66.0247 & 92.43$,$ 279.209  & \textbf{93.9}$,$ 256.628   \\
\hline
Average AUC(\%), Average time(s)     & 77.77$,$ 14.81607 & 77.21$,$ 0.02299 & 77.37$,$ 28.24583 & 77.21$,$ 22.1287 & 89.82$,$ 83.89032 & \textbf{90.84}$,$ 77.65693 \\
\hline

\hline
\end{tabular}}
\end{table*}

\section{Conclusions}
This study proposes novel and enhanced versions of the RELS-TSVM model using fuzzy membership values. To deal with the class imbalance and noise simultaneously, we presented a projection based fuzzy membership assignment (PFMA), and used in our proposed robust fuzzy energy-based least square twin support vector machine (F-RELSTSVM). The proposed algorithm deals with noise present near and away from the hyperplane by using the fusion of energy and projection based approaches. The proposed F-RELSTSVM also involves the 2-norm of the slack variable to ensure strong convexity in the optimization problem.

\par Moreover, we also propose another enhanced version of RELS-TSVM, by using a well-known fuzzy membership technique i.e. intuitionistic fuzzy membership to propose IF-RELSTSVM. This ensures that the optimization problems in IF-RELSTSVM remain insensitive to noise and outliers. Based on the experimental results, it can be concluded that the proposed F-RELSTSVM demonstrates strong generalization performance, particularly in noisy data, surpassing the baseline algorithms. Furthermore, we showed the application of the proposed F-RELSTSVM on credit card fraud detection, where the proposed F-RELSTSVM outperformed all the other baseline algorithms. This justifies the applicability of the proposed algorithm in real world scenarios.

\par In future, we aim to improve the selection process for getting the proximal plane in the proposed fuzzy membership degree by exploring heuristic-based approaches that will reduce training time. Extending our approach to handle multi-class classification problems is an exciting avenue for further investigation.

\bibliography{refs.bib}
\bibliographystyle{unsrtnat}
\end{document}